\documentclass[10pt,twocolumn,letterpaper]{article}

\usepackage{cvpr}              
\definecolor{cvprblue}{rgb}{0.21,0.49,0.74}
\usepackage[pagebackref,breaklinks,colorlinks,allcolors=cvprblue]{hyperref}

\usepackage{amsmath}
\usepackage{multirow} 
\usepackage{graphicx} 
\usepackage{amsfonts} 
\usepackage{booktabs}  
\usepackage{xcolor}
\usepackage{enumitem} 

\def\paperID{810} 
\def\confName{CVPR}
\def\confYear{2026}

\title{CoLC: Communication-Efficient Collaborative Perception \\ with LiDAR Completion}

\author{Yushan~Han$^{1,2}$ \quad Hui~Zhang$^{1,2}$ \quad Qiming~Xia$^{3}$ \quad Yi~Jin$^{1,2}$\quad Yidong~Li$^{1,2}$\thanks{Corresponding author (ydli@bjtu.edu.cn)} \\
$^{1}${Key Laboratory of Big Data \& Artificial Intelligence in Transportation, Ministry of Education} \\ \quad $^{2}${School of Computer Science and Technology, Beijing Jiaotong University} \quad $^{3}${Xiamen University} \\ 
}

\begin{document}
\maketitle
\begin{abstract}

Collaborative perception empowers autonomous agents to share complementary information and overcome perception limitations. 
While early fusion offers more perceptual complementarity and is inherently robust to model heterogeneity, its high communication cost has limited its practical deployment, prompting most existing works to favor intermediate or late fusion. 
To address this, we propose a communication-efficient early \underline{Co}llaborative perception framework that incorporates \underline{L}iDAR \underline{C}ompletion to restore scene completeness under sparse transmission, dubbed as~\texttt{CoLC}.
Specifically, the CoLC integrates three complementary designs. First, each neighbor agent applies Foreground-Aware Point Sampling (FAPS) to selectively transmit informative points that retain essential structural and contextual cues under bandwidth constraints. The ego agent then employs Completion-Enhanced Early Fusion (CEEF) to reconstruct dense pillars from the received sparse inputs and adaptively fuse them with its own observations, thereby restoring spatial completeness. Finally, the Dense-Guided Dual Alignment (DGDA) strategy enforces semantic and geometric consistency between the enhanced and dense pillars during training, ensuring consistent and robust feature learning. 
Experiments on both simulated and real-world datasets demonstrate that CoLC achieves superior perception-communication trade-offs and remains robust under heterogeneous model settings. 

\end{abstract}


\section{Introduction}

Collaborative perception allows multiple agents to share complementary information, fundamentally addressing the limitations of single-agent perception, such as limited perception range, sensor blind spots, and occlusion. 
It has become a key technology in autonomous systems, with significant potential for vehicle-to-everything (V2X) communication-aided autonomous driving \cite{fu2025generative, yuan2025sparsealign,trafalign,polyinter}.

\begin{figure}[htbp]
  \centerline{\includegraphics[width=1\linewidth]{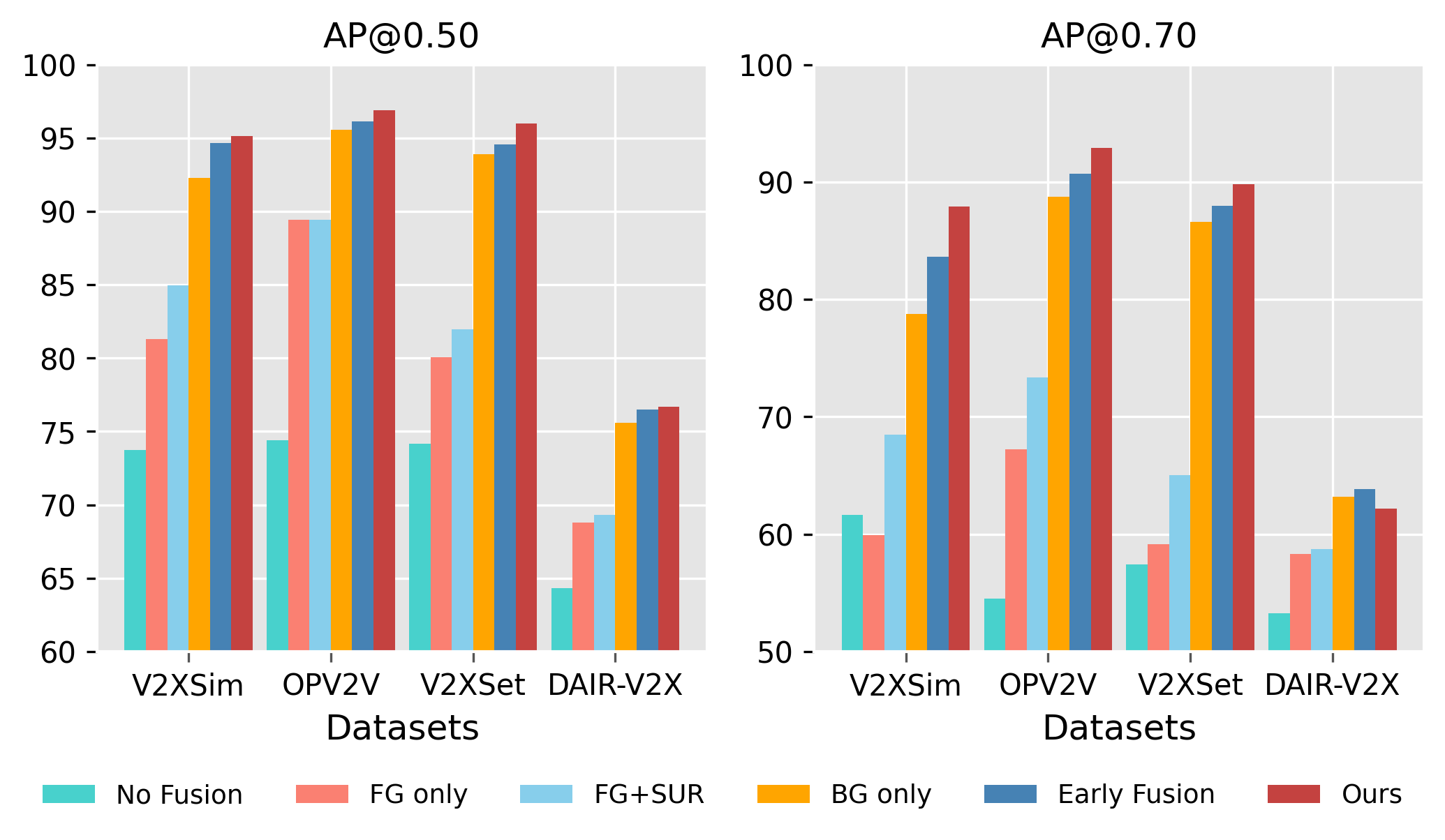}}
  \caption{Comparative performance of early fusion when transmitting foreground (FG) points, foreground with surrounding (FG+SUR) points, background (BG) points, or all points. Transmitting FG or FG+SUR points leads to significant degradation due to missing contextual cues, demonstrating the necessity of incorporating both FG and BG points in early fusion.}
  \label{fig1}
  \vspace{-1.8em}
\end{figure}

In collaborative perception, a key challenge is optimizing performance under bandwidth constraints. Existing methods can be categorized by the type of transmitted information: early fusion with raw point clouds, intermediate fusion with bird's eye view (BEV) features, and late fusion with detection outputs.
Among these, early fusion offers two notable advantages:
(1) Information Preservation: Unlike intermediate and late fusion, which rely on processed features or outputs and thus lose fine-grained geometric details, early fusion \cite{cooper,arnold2020cooperative} directly transmits raw sensor data, preserving complete scene information.
(2) Robustness to Model Heterogeneity: Intermediate and late fusion depend on model-specific representations, making them vulnerable to semantic inconsistencies when agents employ heterogeneous backbones \cite{xu2023bridging,xu2023model,pnpda,HEAL}. In contrast, early fusion is model-agnostic and inherently robust to such discrepancies, enabling more stable collaboration in heterogeneous systems.
However, the high communication cost of early fusion remains a major bottleneck, prompting existing works to adopt intermediate or late fusion as a trade-off. This motivates us to explore strategies that retain the advantages of early fusion while reducing bandwidth consumption.

Due to the critical role of foreground information in collaborative perception, many intermediate fusion approaches \cite{where2comm,codefilling,pointcluster,xu2025cosdh} transmit only foreground-region features while maintaining competitive, or even superior performance.
Motivated by these methods, an intuitive solution is to transmit only foreground point clouds, which directly supplement the observation of the ego while significantly reducing bandwidth consumption. However, transmitting only foreground points results in a notable drop in detection performance compared to early fusion, as shown in Figure~\ref{fig1}.
Adding surrounding neighbor points improves performance over foreground-only transmission, but still underperforms background-only transmission.
In contrast, transmitting only background points yields better performance than both foreground-only and foreground-with-surrounding transmission, albeit still inferior to early fusion. This suggests that both foreground and background points are essential in early fusion: foreground points focus on completing object shapes, while background points serve as contextual cues for object identification and spatial alignment across agents.

\begin{figure*}[htbp]
    \centerline{\includegraphics[width=1\linewidth]{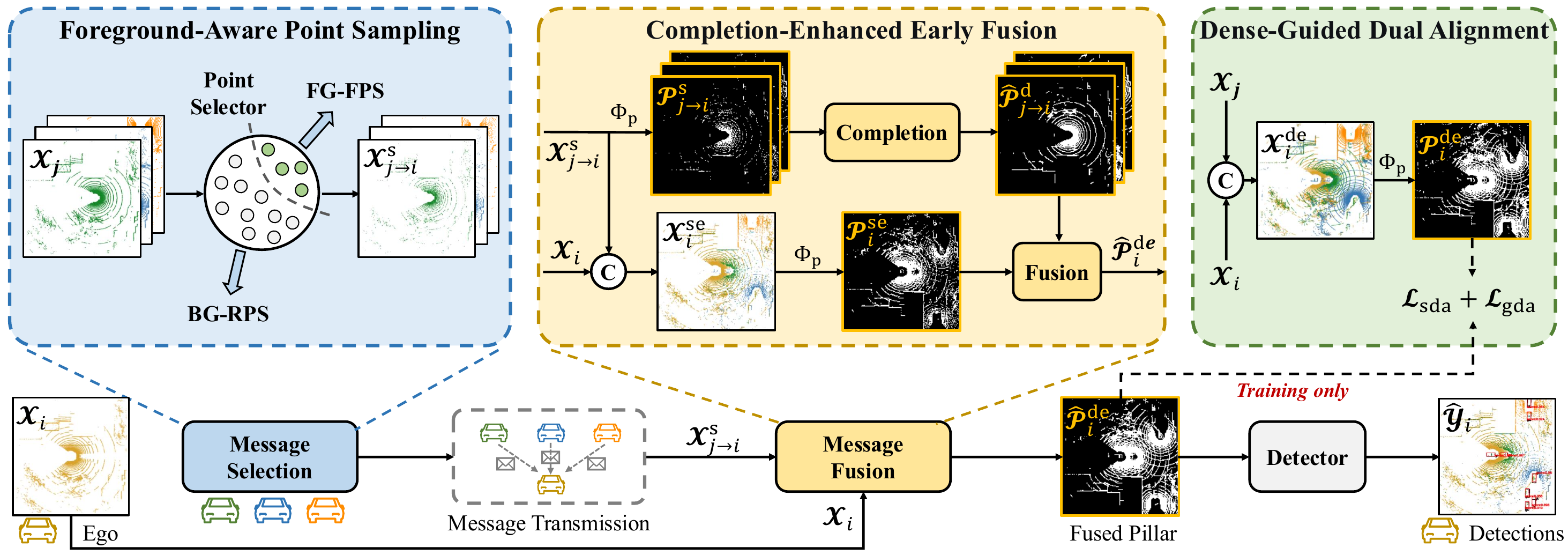}}
    \caption{
    The overall architecture of CoLC. (a) Each neighbor agent applies Foreground-Aware Point Sampling (FAPS) to select and transmit informative points to the ego agent for efficient communication. (b) The ego agent performs Completion-Enhanced Early Fusion (CEEF), where the received sparse points are pillarized $\Phi_{\text{p}}(\cdot)$, completed into dense pillars, and adaptively fused with the initial sparse fusion to enhance structural detail. (c) During training, Dense-Guided Dual Alignment (DGDA) enforces semantic distribution and geometric direction consistency, further improving the alignment of the fused representation.}
    \label{fig2}
    \vspace{-1.8em}
\end{figure*}

Building upon the above insights, we propose CoLC, a communication-efficient early collaborative perception framework. The core idea is to perform spatially-aware point cloud sampling at neighbor agents and apply LiDAR completion at the ego agent, thereby reducing communication cost while reconstructing critical scene information from sparse inputs.
As illustrated in Figure~\ref{fig2}, CoLC consists of three key components: (i) To efficiently convey sparse keypoint information under bandwidth constraints,  each neighbor agent applies Foreground-Aware Point Sampling (FAPS), which adopts distinct sampling strategies to foreground and background regions to reduce communication cost while preserving critical structural and contextual cues.  (ii) After receiving these sparse points, the ego agent employs the Completion-Enhanced Early Fusion (CEEF), which reconstructs dense pillar representations from the received sparse points and adaptively fuses them with the ego observations, thereby enhancing spatial completeness and structural fidelity.  (iii) Finally, the Dense-Guided Dual Alignment (DGDA) further refines the fused representation by aligning the enhanced early-fusion pillars with their dense counterparts in both semantic and geometric spaces, ensuring consistent feature learning during training.

In summary, our contributions are as follows:


\begin{itemize}[itemsep=0pt,topsep=3pt,parsep=5pt]
    \item We propose CoLC, a novel communication-efficient early fusion framework that significantly improves the perception-communication trade-off by combining point cloud sampling with pillar-level LiDAR completion.
    \item The proposed FAPS module preserves structural and contextual information by applying distinct sampling strategies to foreground and background points.
    \item The CEEF module reconstructs dense pillars from sparse neighbor points and fuses them with ego observations to enhance spatial completeness. The DGDA module further guides semantic and geometric consistency between the enhanced early-fusion pillars and dense counterparts.
    \item Extensive experiments demonstrate that CoLC maintains a superior trade-off in terms of the perception performance and communication bandwidth. In addition, CoLC is model-agnostic and naturally robust to heterogeneous model settings across collaborative agents.
\end{itemize}

\section{Related Work}

\textbf{Collaborative Perception.}
Collaborative perception \cite{han2023collaborative,CRCNet,xia2026dota,han2023ssc3od,han2025codts,han2025cods} enhances individual perception by enabling multi-agent information exchange, allowing each agent to access a more comprehensive and holistic view of the environment. 
Due to bandwidth constraints in real-world deployments, prior work has largely focused on intermediate or late fusion, exploring whom to communicate with \cite{who2com}, when to communicate \cite{when2com}, which feature region to transmit \cite{where2comm,codefilling,xu2025cosdh}, and how to compress features along spatial or channel dimensions \cite{core,ERMVP,MRCNet,yang2023how2comm}. While early fusion frameworks \cite{cooper,arnold2020cooperative} preserve raw sensory inputs and are inherently robust to model heterogeneity, this paradigm remains relatively underexplored due to its high communication cost. 
To fill this gap, we propose an efficient early fusion framework that combines spatial-aware point sampling and leverages LiDAR completion, thereby reducing communication volume and restoring critical information.

\noindent
\textbf{LiDAR Completion.}
Training perception models with dense LiDAR is essential for building robust autonomous driving systems. However, dense LiDAR are prohibitively expensive, while low-beam LiDAR produces sparse point clouds that degrade perception accuracy and driving safety.
Recent studies address this issue through LiDAR completion using generative models such as VAEs, GANs and diffusion models \cite{LiDARVAE, ProjGAN, ultralidar, olidm}, aiming to reconstruct dense 3D scenes from sparse inputs.
Nevertheless, these methods are primarily designed for offline scene synthesis and rely on high-capacity models with high-resolution inputs, making them unsuitable for real-time deployment in autonomous vehicles.
STAR \cite{li2023star} further introduces a MAE-based reconstruction for masked features in intermediate collaboration, yet such masking strategies are not directly applicable to early collaboration.
In this work, we innovatively introduce LiDAR completion into early collaborative perception to reach a trade-off between perception performance and communication cost. 
To this end, we adopt a pillar-level vector quantization (VQ) \cite{vqvae} -based LiDAR completion module, which enables compact and memory-efficient encoding while preserving essential structural details in the reconstructed representations.

\section{Problem Formulation}
Consider $N$ agents in the scene. let $\mathcal{X}_i$ and $\mathcal{Y}_i$ represent the LiDAR data and supervision of the $i$-th agent, respectively. The objective of early collaborative perception is to achieve the maximized perception performance of all agents as a function of the total communication budget $B$:
\begin{equation}
    \begin{aligned}
    &\max _{\theta, \mathcal{X}} \sum_i g\Big(\Phi_\theta\big(\mathcal{X}_i, \{\mathcal{M}_{j \rightarrow i}\}_{j=1, j\neq i}^N\big), \mathcal{Y}_i\Big),\\
    &\text{s.t. } \sum_{i, j, j \neq i} b\left(\mathcal{M}_{j \rightarrow i}\right) \leq B,
    \end{aligned}
    \end{equation}
where $g(\cdot, \cdot)$ is the detection evaluation metric, $\Phi(\cdot)$ is a LiDAR-based 3D detector with trainable parameter $\theta$, $\mathcal{M}_{j \rightarrow i}$ is the message transmitted from the $j$-th agent to the $i$-th agent, and $b(\cdot)$ measures the communication cost of $\mathcal{M}_{j \rightarrow i}$. In early collaborative perception, $\mathcal{M}_{j \rightarrow i}$ denotes the full point cloud $\mathcal{X}_j$ observed by the $j$-th agent. We aim to reduce the transmission overhead of $\mathcal{M}_{j \rightarrow i}$ while preserving high perception performance.

\section{Method}

The overall architecture of the CoLC is shown in Figure~\ref{fig2}.
(i) At neighbor agents, FAPS applies a spatially-aware sampling strategy to selectively transmit informative points, effectively reducing communication overhead while preserving essential structural cues.
(ii) At the ego agent, CEEF reconstructs dense pillars from the received sparse inputs, providing structural refinement to the initial sparse fusion results.
(iii) DGDA further enhances the fused pillar by enforcing semantic and geometric consistency during training.

\subsection{Foreground-Aware Point Sampling}
In intermediate and late fusion, semantic information has already been extracted from raw point clouds, allowing the model to focus on foreground complementarity and fusion. In contrast, early fusion requires the detectors to learn features directly from both the ego and the received point clouds. Specifically, foreground points contribute to object shape completion, while background points act as geometric anchors that ensure coherent spatial alignment between agents. Motivated by this, we propose a foreground-aware point sampling strategy that considers both foreground and background regions to preserve structural and contextual information under communication constraints.

For the $j$-th agent, given its original point cloud $\mathcal{X}_j \in \mathbb{R}^{M \times 4}$, we first employ an independently pre-trained, lightweight MLP-based point-wise selector to estimate a saliency map $\mathcal{S}_j \in [0,1]^{M}$. We then partition $\mathcal{X}_j$ into foreground and background sets via a threshold $\tau_s$: $\mathcal{X}_j^{fg} = \{ \mathbf{x}_i \mid \mathcal{S}_j^i > \tau_s \}$ and $\mathcal{X}_j^{bg} = \{ \mathbf{x}_i \mid \mathcal{S}_j^i \leq \tau_s \}$.

Subsequently, we apply distinct sampling strategies to the two point sets:
(i) \textbf{Foreground Farthest Point Sampling (FG-FPS)}:
Foreground points typically form compact object structures. To preserve this structural integrity, we apply FPS \cite{pointnet++} with a ratio $R^{fg}$ to $\mathcal{X}_j^{fg}$. Given the typically small number of foreground points, FPS introduces negligible inference overhead.
(ii) \textbf{Background Random Point Sampling (BG-RPS)}:
In contrast, the background point set $\mathcal{X}_j^{bg}$ is substantially larger, making FPS computationally expensive. Therefore, we adopt RPS with a ratio $R^{bg}$ to obtain a sparse subset of background points efficiently.
The final transmitted message is a sparse point cloud $\mathcal{X}_{j \rightarrow i}^{\text{s}}$, which contains the sampled foreground and background points.

\begin{figure}[htbp]
    \centerline{\includegraphics[width=1\linewidth]{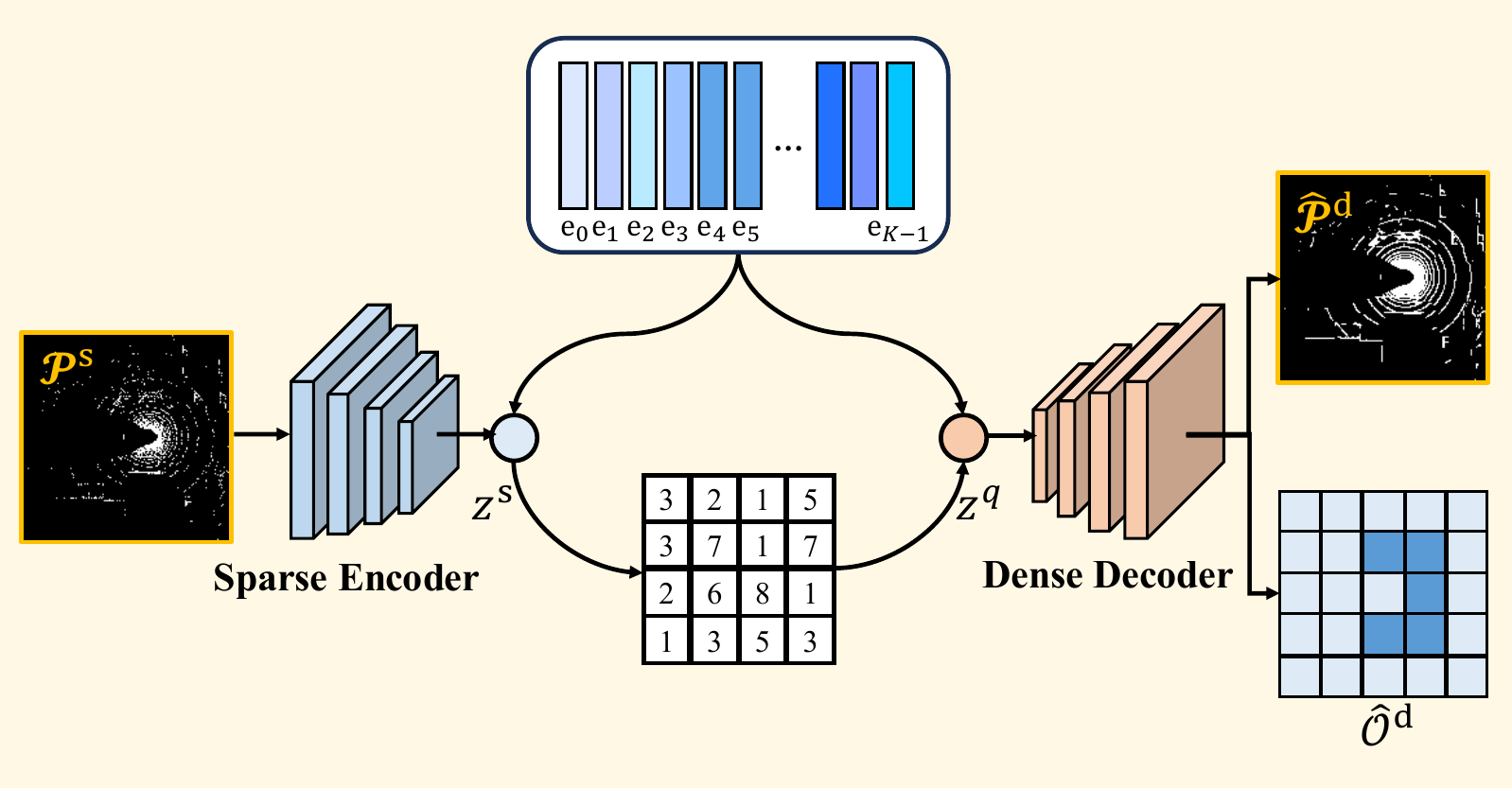}}
    \caption{VQ-based LiDAR completion.}
    \label{fig3}
    \vspace{-1em}
\end{figure}

\subsection{Completion-Enhanced Early Fusion}

To compensate for information loss caused by point sampling, we introduce a novel VQ-based pillar-level LiDAR completion module. 
It uses compact codebook embeddings to enable structurally consistent pillar reconstruction. Compared to prior works \cite{nunes2024scaling, olidm} that rely on high-resolution inputs and large-capacity generative models, our approach reduces memory and computational demands while maintaining sufficient structural fidelity for downstream detection. Although the module operates on pillar-level representations \cite{pointpillars}, it is conceptually compatible with voxel-based 3D object detectors \cite{second, voxelnet}, ensuring seamless integration into existing perception pipelines.

Following the pre-training of the completion module, the CEEF performs progressive fusion in three stages: (i) initial sparse early fusion, (ii) parallel pillar completion for each neighbor, and (iii)  adaptive complementary fusion that refines the initial fusion result with completed pillars.

\subsubsection{VQ-based LiDAR Completion}

Given a pair of sparse and dense (full) point clouds ${\mathcal{X}^{\text{s}}, \mathcal{X}^{\text{d}}}$ from the same agent, we first apply pillarization $\Phi_{\text{p}}(\cdot)$ to obtain their corresponding pillars: $\mathcal{P}^{\text{s}} = \Phi_{\text{p}}(\mathcal{X}^{\text{s}})$ and $\mathcal{P}^{\text{d}} = \Phi_{\text{p}}(\mathcal{X}^{\text{d}})\in\mathbb{R}^{H \times W \times C}$, where $H, W, C$ are height, width and channel, respectively. 
To reconstruct dense pillars $\mathcal{P}^\text{d}$ from sparse inputs $\mathcal{P}^\text{s}$, the VQ-based completion module $\Psi(\cdot)$ consists of three components, as shown in Figure~\ref{fig3}:

(1) \textbf{Sparse Encoder.}
The sparse pillar $\mathcal{P}^{\text{s}}$ encodes low-level occupancy and geometric cues, which are insufficient for reconstructing dense structures.
To obtain globally contextualized BEV representations, the sparse encoder $\Psi_{\text{enc}}^{\text{s}}$ adopts a Swin Transformer \cite{liu2021swin} (depth $L$, embedding dim $D$) to extract $P$ patch embeddings. The features are then projected into the quantization space $D_c$ as $\mathbf{z}^{\text{s}}\in\mathbb{R}^{P\times D_c}$.

\begin{figure}[htbp]
    \centerline{\includegraphics[width=1\linewidth]{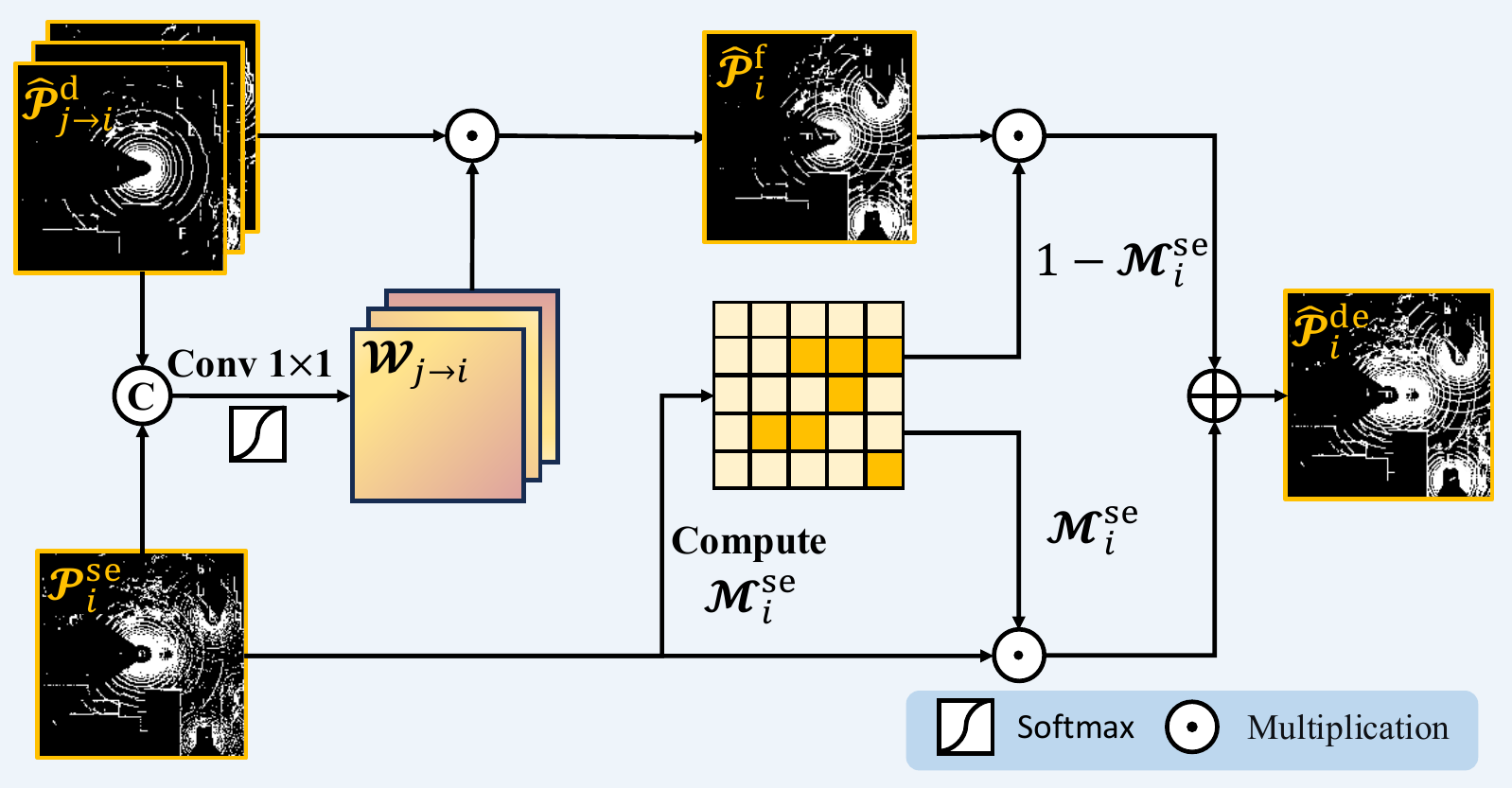}}
    \caption{Adaptive complementary fusion.
      }
    \label{fig4}
    \vspace{-1em}
\end{figure}

(2) \textbf{Vector Quantization.} 
We maintain a learnable codebook $E = \{\mathbf{e}_k\}_{k=0}^{K-1} \in \mathbb{R}^{K \times D_c}$, where $K$ represents codebook size. The vector quantizer $\Psi_{\text{vq}}$ maps each continuous latent vector $\mathbf{z}_i^{\text{s}}$ to its nearest neighbor in the codebook:
\begin{equation}
\mathbf{z}_i^{\text{q}} = \Psi_{\text{vq}}(\mathbf{z}_i^{\text{s}}; E) = \mathbf{e}_k, \quad \text{where} \quad k = \arg\min_j \|\mathbf{z}_i^{\text{s}} - \mathbf{e}_j\|_2.
\end{equation}

(3) \textbf{Dense Decoder.} The dense decoder $\Psi_{\text{dec}}^{\text{d}}$ maps the quantized embeddings $\mathbf{z}^{\text{q}}$ back to the pillar space, producing a reconstructed dense pillar $\hat{\mathcal{P}}^{\text{d}} \in \mathbb{R}^{H \times W \times C}$.
In addition, since the pillar contains a large number of empty grids, we further predict pillar-wise occupancy mask $\hat{\mathcal{O}}^{\text{d}}\in \mathbb{R}^{H \times W \times 1}$ via a convolutional layer.

Finally, the completion module is trained with a reconstruction loss $\mathcal{L}_\text{rec}$ and a vector quantization loss $\mathcal{L}_\text{vq}$.
To focus reconstruction on occupied regions, we define the target binary occupancy mask $\mathcal{O}^\text{d} = \left( \sum_{c=1}^{C} |\mathcal{P}^\text{d}_{:,:,c}| > 0 \right) \in \{0,1\}^{H \times W \times 1}$. 
The $\mathcal{L}_\text{rec}$ then comprises: (i) a binary cross-entropy loss $\text{BCE}(\cdot)$ for occupancy prediction and (ii) a mean squared error term over occupied pillars only:
\begin{equation}
    \mathcal{L}_{\text{rec}} = \text{BCE}(\hat{\mathcal{O}}^{\text{d}}, \mathcal{O}^\text{d}) + \frac{1}{|\mathcal{O}^\text{d}|} \sum_{i,j} \mathcal{O}^\text{d}_{i,j} \|\hat{\mathcal{P}}_{i,j}^{\text{d}} - \mathcal{P}_{i,j}^{\text{d}}\|_2^2.
\end{equation}


The vector quantization loss $\mathcal{L}_{\text{vq}}$ is used to jointly train the codebook and the encoder.
Formally, it consists of a codebook loss and a commitment loss, defined as:
\begin{equation}
    \mathcal{L}_{\text{vq}} = \left\| \mathrm{sg}[\mathbf{z}^{\text{s}}] - \mathbf{z}^\text{q} \right\|_2^2 + \beta \left\| \mathbf{z}^{\text{s}} - \mathrm{sg}[\mathbf{z}^\text{q}] \right\|_2^2 ,
\end{equation}
where $\mathrm{sg}[\cdot]$ is the stop-gradient operator \cite{vqvae} and $\beta$ controls the commitment weight. 
The overall loss is defined as:
\begin{equation}
    \mathcal{L}_\Psi = \lambda \cdot \mathcal{L}_{\text{rec}} + \mathcal{L}_{\text{vq}},
\end{equation}
where $\lambda$ is a weighting factor that balances $\mathcal{L}_{\text{rec}}$ and $\mathcal{L}_{\text{vq}}$.



\subsubsection{Adaptive Complementary Fusion}

After training the LiDAR completion module, we freeze it and implement early fusion as follows. 
Specifically, the $i$-th agent receives sparse point clouds $\left\{\mathcal{X}_{j \rightarrow i}^{\text{s}}\right\}_{j=1, j\neq i}^N$. It then concatenates its own point cloud $\mathcal{X}_i$ with the received ones to form $\mathcal{X}_i^{\text{se}} = [\mathcal{X}_i | \left\{\mathcal{X}_{j \rightarrow i}^{\text{s}}\right\}_{j=1, j\neq i}^N]$, where $[\cdot|\cdot]$ denotes concatenation operation. This aggregated point cloud is subsequently pillarized as $\mathcal{P}_i^{\text{se}} = \Phi_{\text{p}}(\mathcal{X}_i^{\text{se}})$.

For each sparse point cloud $\mathcal{X}_{j \rightarrow i}^{\text{s}}$, we first obtain its pillar via $\mathcal{P}_{j \rightarrow i}^{\text{s}} = \Phi_{\text{p}}(\mathcal{X}_{j \rightarrow i}^{\text{s}})$. 
Then we complete pillars via:
\begin{equation}
\hat{\mathcal{P}}_{j \rightarrow i}^{\text{d}}, \hat{\mathcal{O}}_{j \rightarrow i}^{\text{d}}  = \Psi_{\text{dec}}^{\text{d}}(\Psi_{\text{vq}}(\Psi_{\text{enc}}^{\text{s}}(\mathcal{P}_{j \rightarrow i}^{\text{s}}),E)).
\end{equation}

To suppress unreliable completions in empty regions, we retain only those pillars whose predicted occupancy $\hat{\mathcal{O}}_{j \rightarrow i}^{\text{d}}$ exceeds a threshold $\tau_o$. To further preserve the fidelity of the original sparse observations, we replace the corresponding locations in the completed pillars $\hat{\mathcal{P}}_{j \rightarrow i}^{\text{d}}$ with values from the sparse pillars $\mathcal{P}_{j \rightarrow i}^{\text{s}}$.
This hybrid scheme ensures structural enhancement while maintaining geometric consistency.

Subsequently, we adopt an adaptive complementary fusion strategy to refine the sparse early fusion pillar $\mathcal{P}_i^{\text{se}}$ using the completed neighbor pillars $\hat{\mathcal{P}}_{j \rightarrow i}^{\text{d}}$. 
As shown in Figure~\ref{fig4}, we compute a spatial correlation map $\mathcal{W}_{j \rightarrow i}$ by concatenating $\mathcal{P}_i^{\text{se}}$ and $\hat{\mathcal{P}}_{j \rightarrow i}^{\text{d}}$ along the channel dimension, followed by several $1\times1$ convolutional layers and a softmax operation to produce normalized fusion weights. 
The fused complementary feature from neighbor agents is computed as:
\begin{equation}
\hat{\mathcal{P}}_{i}^{\text{f}} = \sum_{j=1,\, j\neq i}^{N} \mathcal{W}_{j \rightarrow i} \odot \hat{\mathcal{P}}_{j \rightarrow i}^{\text{d}},
\end{equation}
where $\odot$ denotes element-wise multiplication.

To preserve the fidelity of the observed regions, we generate an occupancy mask for $\mathcal{P}_i^{\text{se}}$ as 
$\mathcal{M}_{i}^{\text{se}} = \left( \sum_{c=1}^{C} \left| \mathcal{P}_i^{\text{se}}[:, :, c] \right| > 0 \right) \in \{0, 1\}^{H \times W \times 1}$, 
indicating the occupied pillar locations in the early fusion result. 
We then selectively update only the empty pillars using the fused features, yielding the final enhanced early-fusion pillar representation $\hat{\mathcal{P}}_i^{\text{de}}$ for downstream detection:
\begin{equation}
\hat{\mathcal{P}}_i^{\text{de}} = 
\mathcal{M}_{i}^{\text{se}} \odot \mathcal{P}_i^{\text{se}} + 
(1 - \mathcal{M}_{i}^{\text{se}}) \odot \hat{\mathcal{P}}_{i}^{\text{f}}.
\end{equation}

\subsection{Dense-Guided Dual Alignment}

Although LiDAR completion improves spatial coverage, the enhanced early-fusion pillars may still exhibit noisy artifacts and structural inconsistencies compared to their dense counterparts.
To address this, we introduce DGDA, which aligns the enhanced early-fusion pillars with their dense counterparts in both semantic distribution and geometric direction spaces.
This dual alignment is motivated by the dual nature of pillar features, which encode partial semantic abstraction while preserving explicit geometric structure, necessitating joint supervision for consistent alignment.

Specifically, for the $i$-th agent, we transmit full point clouds $\left\{\mathcal{X}_{j}\right\}_{j=1, j\neq i}^N$ from neighbor agents to construct a dense early fusion point cloud $\mathcal{X}_i^{\text{de}} = [\mathcal{X}_i | \left\{\mathcal{X}_{j}\right\}_{j=1, j\neq i}^N]$, which is then pillarized into $\mathcal{P}_i^{\text{de}}$. 
During training, the enhanced early-fusion pillar $\hat{\mathcal{P}}_i^{\text{de}}$ is encouraged to align with its dense counterpart $\mathcal{P}_i^{\text{de}}$ in both semantic and geometric spaces, through two complementary objectives:

(1) \textbf{Semantic Distribution Alignment.} To reduce semantic discrepancies, we align the channel-wise distributions of $\hat{\mathcal{P}}_i^{\text{de}}$ and $\mathcal{P}_i^{\text{de}}$ using Kullback-Leibler divergence:
\begin{equation}
    \mathcal{L}_{\text{sda}} = D_{\mathrm{KL}}\left(\sigma\left(\hat{\mathcal{P}}_i^{\text{de}}\right) || \sigma\left(\mathcal{P}_i^{\text{de}}\right)\right),
\end{equation}
where $\sigma(\cdot)$ denotes the softmax operation applied along the channel dimension.

(2) \textbf{Geometric Direction Alignment.}
To preserve structural consistency, we encourage directional similarity between $\hat{\mathcal{P}}_i^{\text{de}}$ and $\mathcal{P}_i^{\text{de}}$ via marginal cosine similarity:
\begin{equation}
    \mathcal{L}_{\text{gda}} = \mathbb{E}_i \left[ 1 - \frac{\hat{\mathcal{P}}_i^{\text{de}} \cdot \mathcal{P}_i^{\text{de}}}{\|\hat{\mathcal{P}}_i^{\text{de}}\| \|\mathcal{P}_i^{\text{de}}\|} \right].
\end{equation}

The overall loss $\mathcal{L}_\Phi$ for the detector is defined as:
\begin{equation}
    \mathcal{L}_\Phi = \mathcal{L}_{\text{det}} + \gamma_1 \cdot \mathcal{L}_{\text{sda}} + \gamma_2 \cdot \mathcal{L}_{\text{gda}},
\end{equation}
where $\mathcal{L}_{\text{det}}$ denotes the standard 3D object detection loss \cite{pointpillars}, which adopts focal loss for classiﬁcation, the smooth $L_1$ loss for regression and softmax classification loss for direction. The hyperparameters $\gamma_1$ and $\gamma_2$ controls the weight of the two alignment loss term.

  \begin{table*}[htbp]
    \small
    \caption{Performance comparison (AP@0.3 $\slash$ 0.5 $\slash$ 0.7) on collaborative 3D object detection methods.}
    \label{tab1}
    \centering 
    \resizebox{1\linewidth}{!}{
      \begin{tabular}{c|c|c|c|c}
        \toprule 
        Methods & V2XSim & OPV2V & V2XSet & DAIR-V2X \\ 
        \midrule
        No Fusion  & 74.73 $\slash$ 73.72 $\slash$ 61.65  & 77.65 $\slash$ 74.42 $\slash$ 54.52  & 78.41 $\slash$ 74.18 $\slash$ 57.43  & 67.15 $\slash$ 64.32 $\slash$ 53.27  \\ 
        Late Fusion  & 89.23 $\slash$ 86.87 $\slash$ 78.78  & 94.22 $\slash$ 92.51 $\slash$ 77.82  & 91.81 $\slash$ 89.05 $\slash$ 75.72  & 74.12 $\slash$ 62.05 $\slash$ 42.82  \\ 
        \midrule
        V2VNet \cite{v2vnet}  & 88.16	$\slash$ 87.11	$\slash$ 69.73  & 97.09 $\slash$ 95.89 $\slash$ 83.56  & 92.70 $\slash$ 90.59 $\slash$ 74.52  & 78.15 $\slash$ 70.24 $\slash$ 43.71  \\ 
        DiscoNet \cite{DiscoNet} & 80.73 $\slash$ 77.46 $\slash$ 64.03  & 95.45 $\slash$ 93.45 $\slash$ 82.95  & 89.63 $\slash$ 86.28 $\slash$ 71.03  & 80.32 $\slash$ 73.21 $\slash$ 57.38  \\ 
        AttFusion \cite{OPV2V} & 77.08 $\slash$ 74.51 $\slash$ 64.86  & 94.93 $\slash$ 93.13 $\slash$ 80.66  & 90.07 $\slash$ 87.32 $\slash$ 72.26  & 79.69 $\slash$ 73.46 $\slash$ 58.82  \\ 
        V2X-ViT \cite{V2X-VIT}  & 90.42 $\slash$ 87.77 $\slash$ 70.86  & 96.35 $\slash$ 95.01 $\slash$ 85.84  & 93.93 $\slash$ 90.28 $\slash$ 74.75  & 78.34 $\slash$ 70.62 $\slash$ 54.02  \\ 
        Where2comm \cite{where2comm} & 89.66 $\slash$ 88.45 $\slash$ 80.54  & 95.75 $\slash$ 95.10 $\slash$ 88.48  & 91.98 $\slash$ 90.68 $\slash$ 80.48  & 80.56 $\slash$ 76.70 $\slash$ 61.96  \\ 
        CoBEVT \cite{xu2022cobevt} & 88.04 $\slash$ 85.57 $\slash$ 69.99  & 95.61 $\slash$ 92.48 $\slash$ 71.16  & 91.61 $\slash$ 88.16 $\slash$ 73.08  & 80.22 $\slash$ 74.10 $\slash$ 60.23  \\ 
        ERMVP \cite{ERMVP} & \textbf{95.72} $\slash$ 94.35 $\slash$ 84.76 & 96.80 $\slash$ 95.99 $\slash$ 89.14 & 94.88 $\slash$ 93.08 $\slash$ 81.91 & 79.24 $\slash$ 74.73 $\slash$ 60.75 \\ 
        MRCNet \cite{MRCNet} & 77.64 $\slash$ 75.40 $\slash$ 64.49 & \textbf{97.85} $\slash$ \textbf{96.95} $\slash$ 90.66 & \textbf{97.10} $\slash$ 95.41 $\slash$ 84.32 & 77.87 $\slash$ 73.10 $\slash$ 57.48 \\ 
        CoSDH \cite{xu2025cosdh} &  86.29 $\slash$ 85.33 $\slash$ 77.57  & 96.10 $\slash$ 95.63 $\slash$ 89.82  & 92.70 $\slash$ 91.65 $\slash$ 81.51 &  80.00 $\slash$ 76.62 $\slash$ 60.77 \\ 
        \midrule
        Early Fusion & 95.48 $\slash$ 94.68 $\slash$ 83.61  & 96.77 $\slash$ 96.13 $\slash$ 90.69  & 95.28 $\slash$ 94.59 $\slash$ 88.00  & 80.51 $\slash$ 76.51 $\slash$ \textbf{63.83}  \\ 
        \textbf{CoLC (Ours, 100\% LiDAR)} & 95.63 $\slash$ \textbf{95.14	$\slash$ 87.89} & 97.18 $\slash$ 96.88 $\slash$ \textbf{92.93} & 96.59 $\slash$ \textbf{95.97 $\slash$ 89.81} & \textbf{80.91 $\slash$ 76.71} $\slash$ 62.17 \\
        CoLC* (Ours, 50\% LiDAR) & 94.09 $\slash$ 93.47 $\slash$ 85.28 & 96.94 $\slash$ 96.46 $\slash$ 91.95 & 95.87 $\slash$ 95.05 $\slash$ 87.72  & 80.37 $\slash$ 76.03 $\slash$ 62.09 \\
        \bottomrule
    \end{tabular}
    }
    
    \vspace{-1em}
  \end{table*}

\section{Experiments}
\subsection{Experimental Settings}
\paragraph{Datasets and Evaluation Metrics.} 

We evaluate CoLC on four datasets: V2XSim \cite{DiscoNet}, OPV2V \cite{OPV2V}, V2XSet \cite{V2X-VIT} and DAIR-V2X \cite{yu2022dair} for LiDAR-based 3D object detection,
covering both simulated and real-world scenarios.

For collaborative detection, we evaluate performance using Average Precision (AP) at Intersection-over-Union (IoU) thresholds of 0.3, 0.5, and 0.7. 
The communication volume is measured by the number of transmitted elements $\mathcal{M}$ as $\log _2(\mathcal{M} \times 32 / 8)$, where each element is a 32-bit float (float32), and division by 8 converts bits to bytes \cite{where2comm,codefilling}.

For sparse-to-dense LiDAR completion, we adopt Intersection-over-Union (IoU) \cite{ultralidar} to measure the occupancy prediction accuracy, and Mean Squared Error (MSE) \cite{olidm} to quantify the reconstruction similarity between the completed pillars and the dense ground truth.

\noindent
\textbf{Implementation Details.}
We implement CoLC on the OpenCOOD framework \cite{OPV2V}, using PointPillars \cite{pointpillars} as the detector with a grid size of $(0.4\text{m}, 0.4\text{m})$ and pillar feature channel $C{=}64$. The training is conducted in two stages: first, the LiDAR completion module is pre-trained to convergence using the AdamW optimizer (learning rate 8e-4); subsequently, it is frozen and the entire CoLC pipeline is trained end-to-end with Adam (learning rate 2e-3). 
The point selector in FAPS is pre-trained with point-level supervision, where points inside ground-truth boxes are labeled as foreground.
During training, we use RPS at a 0.1 ratio for point sampling for efficiency, which generalizes well to the FAPS sampling used at inference.
In FAPS, the selection threshold $\tau_s$ is set to $0.5$. In CEEF, $K$, $D_c$, $D$ and $L$ in the completion module are 128, 128, 128, 6. The loss weights $\beta$ and $\lambda$ are set to $0.25$ and $10$, and the occupancy threshold $\tau_o$ to $0.5$. In DGDA, $\gamma_1$ and $\gamma_2$ are set to $1{,}000$ and $10$. All models are trained on NVIDIA RTX 4090 GPUs.
To mitigate point misalignment caused by pose errors and latency, the Iterative Closest Point (ICP) algorithm \cite{zhang2021fast} is applied to CoLC and early fusion baseline (see supplementary).

%

\subsection{Quantitative Evaluation}

\begin{figure*}[htbp]
    \centering
    \begin{minipage}[t]{0.245\textwidth}
        \centering
        \includegraphics[width=\textwidth]{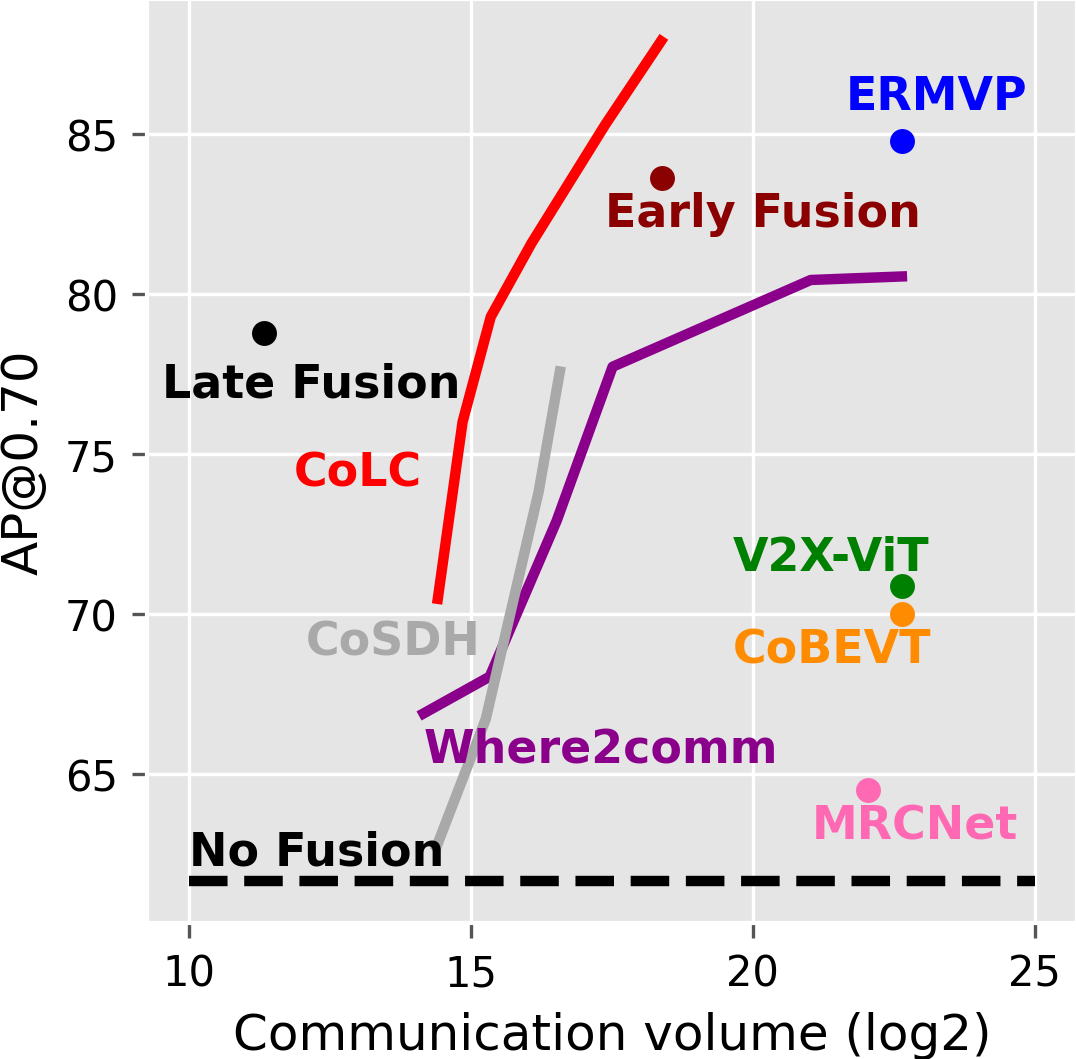}
        \captionsetup{font=small}
        \caption*{(a) Accuracy-Bandwidth.}
    \end{minipage}
    \begin{minipage}[t]{0.245\textwidth}
        \centering
        \includegraphics[width=\textwidth]{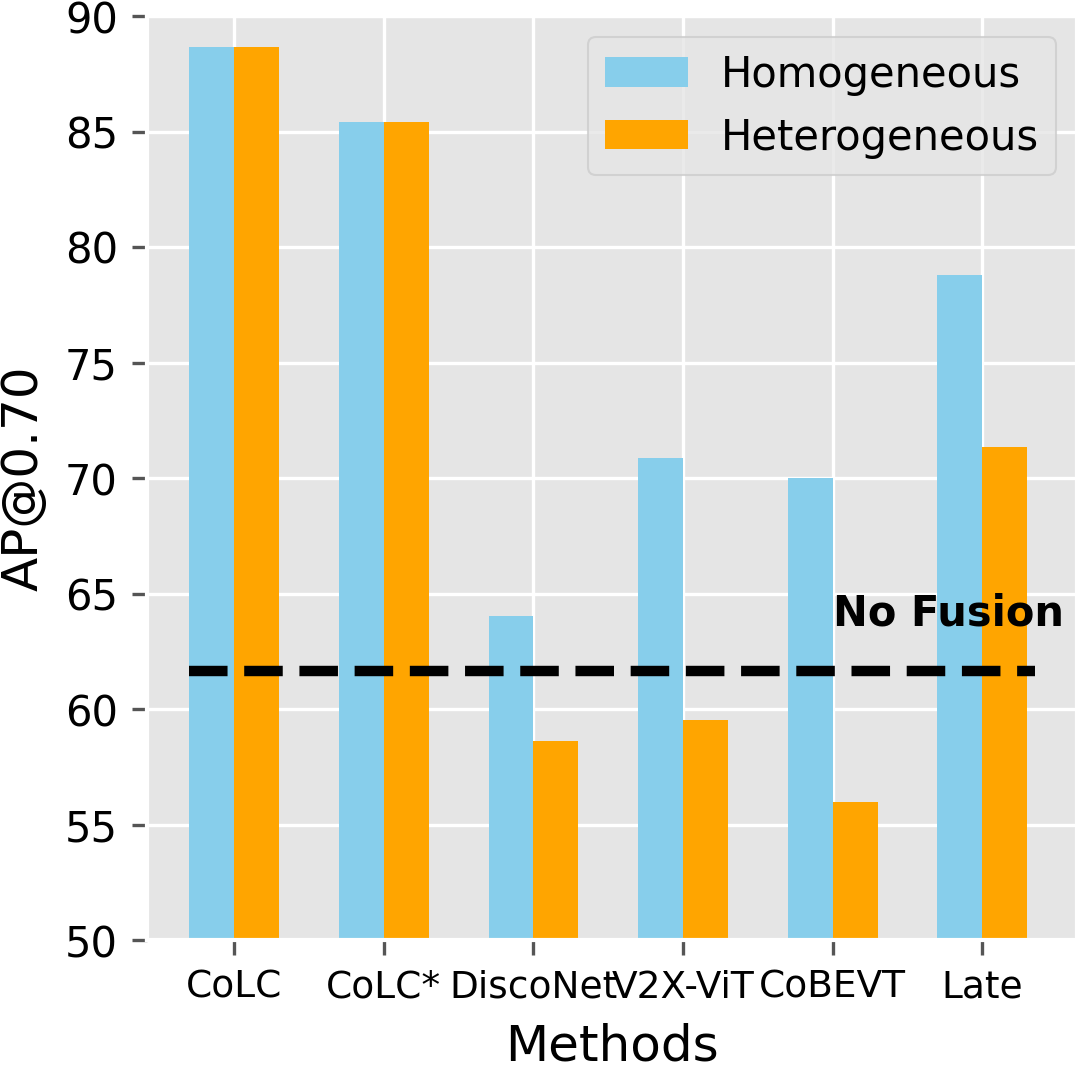}
        \captionsetup{font=small}
        \caption*{(b) Heterogeneous Scenarios.}
    \end{minipage}
    \begin{minipage}[t]{0.245\textwidth}
        \centering
        \includegraphics[width=\textwidth]{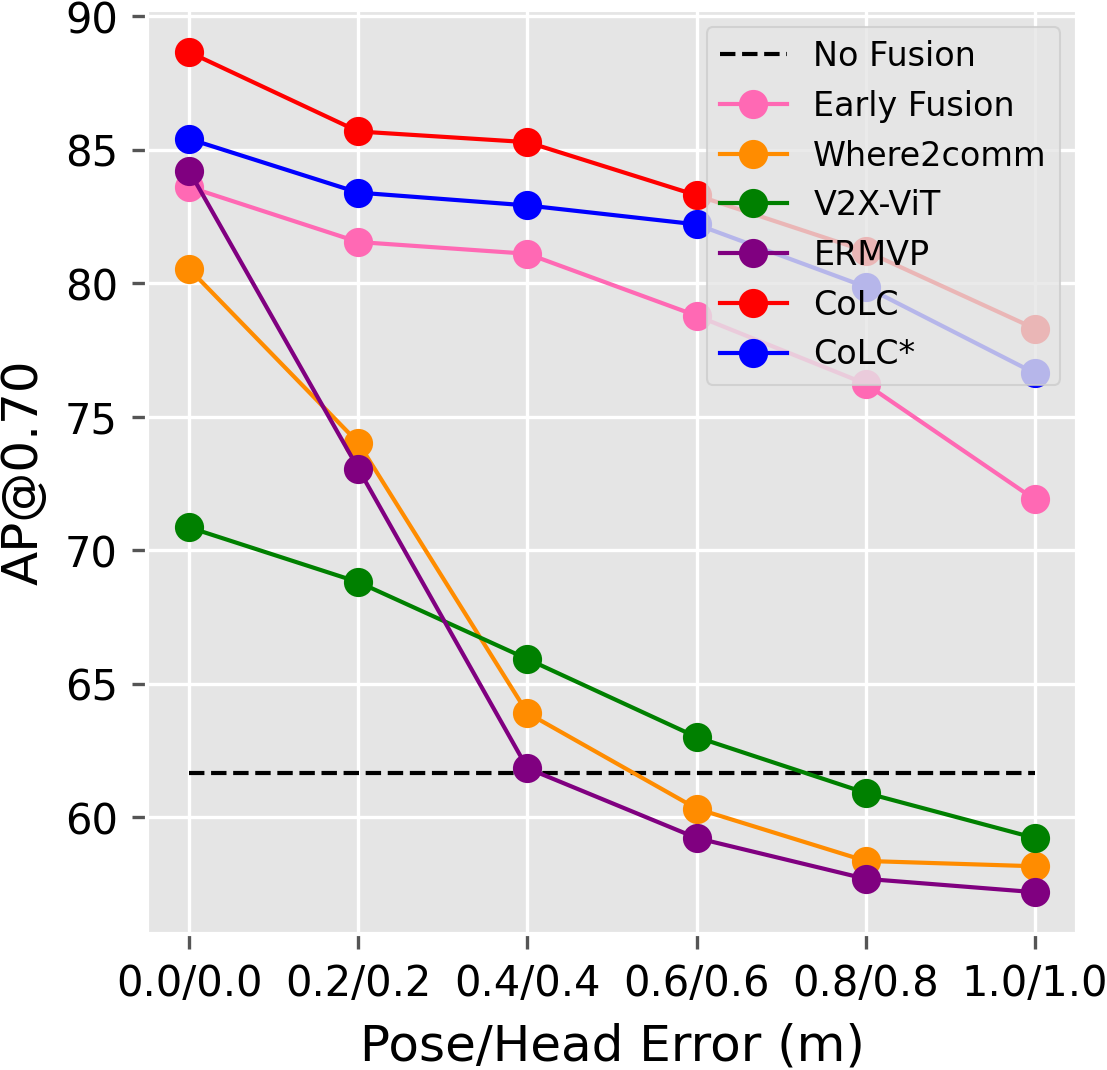}
        \captionsetup{font=small}
        \caption*{(c) Localization Error.}
    \end{minipage}
    \begin{minipage}[t]{0.245\textwidth}
        \centering
        \includegraphics[width=\textwidth]{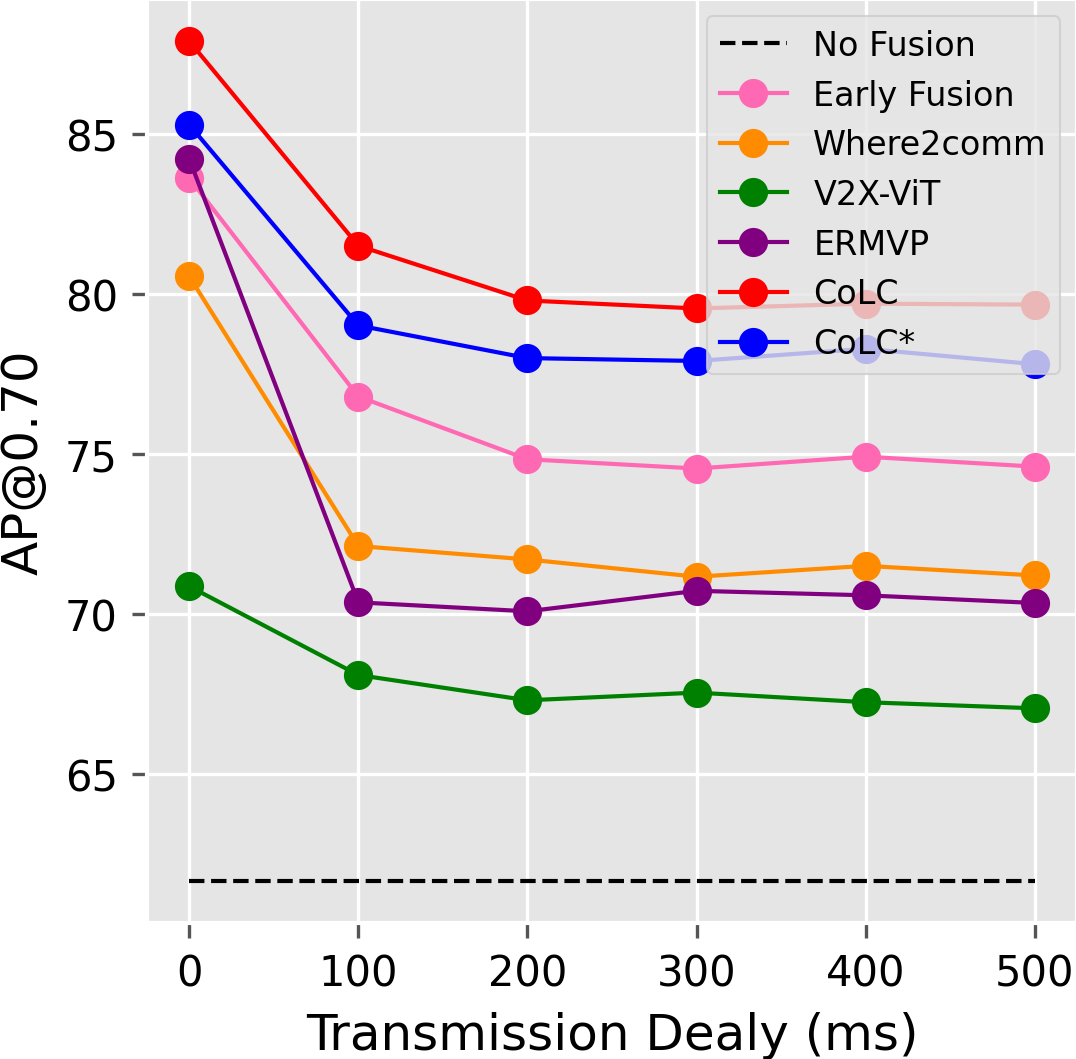}
        \captionsetup{font=small}
        \caption*{(d) Communication Latency.}
    \end{minipage}
    \begin{minipage}[t]{0.245\textwidth}
        \centering
        \includegraphics[width=\textwidth]{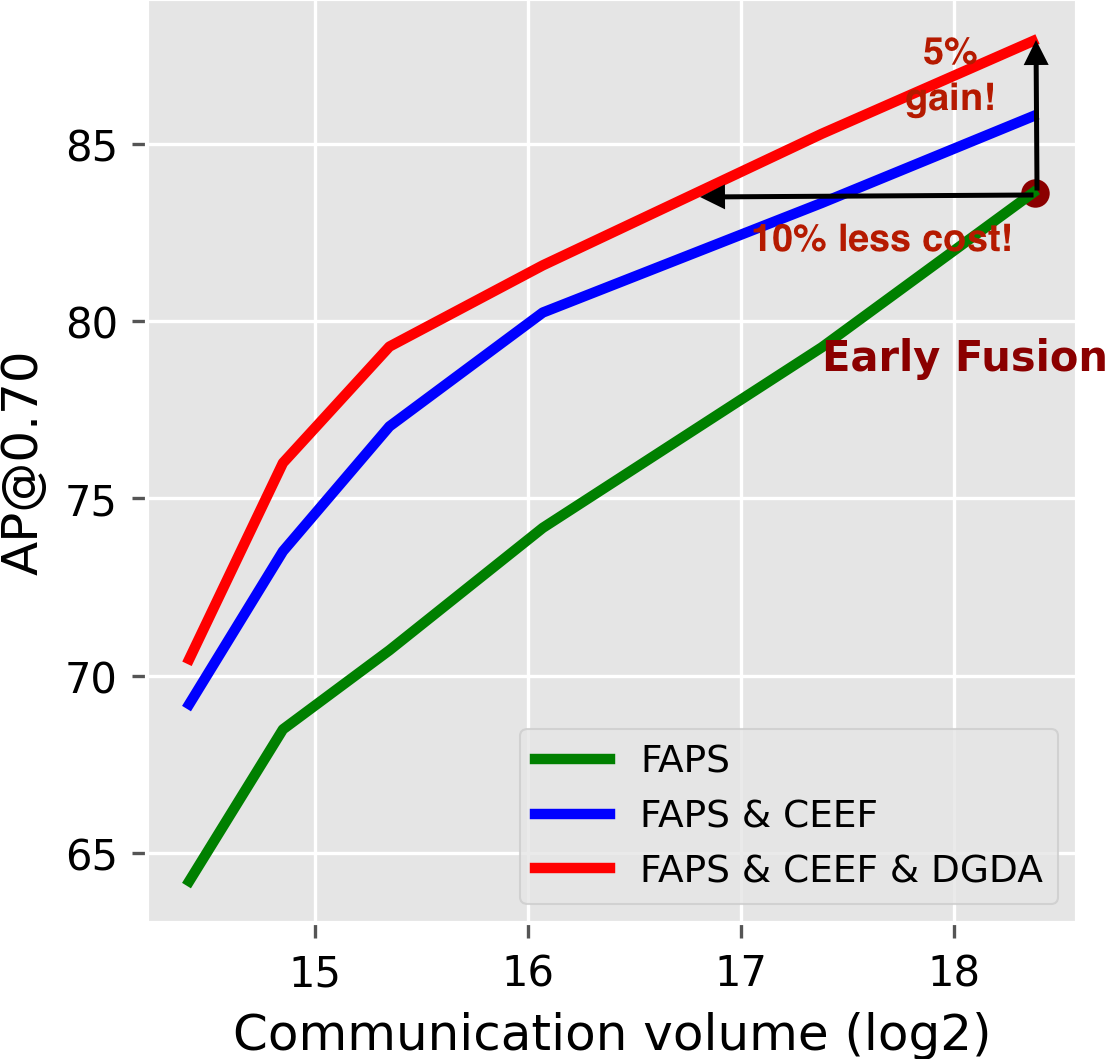}
        \captionsetup{font=small}
        \caption*{(e) Ablation on Components.}
    \end{minipage}
    \begin{minipage}[t]{0.245\textwidth}
        \centering
        \includegraphics[width=\textwidth]{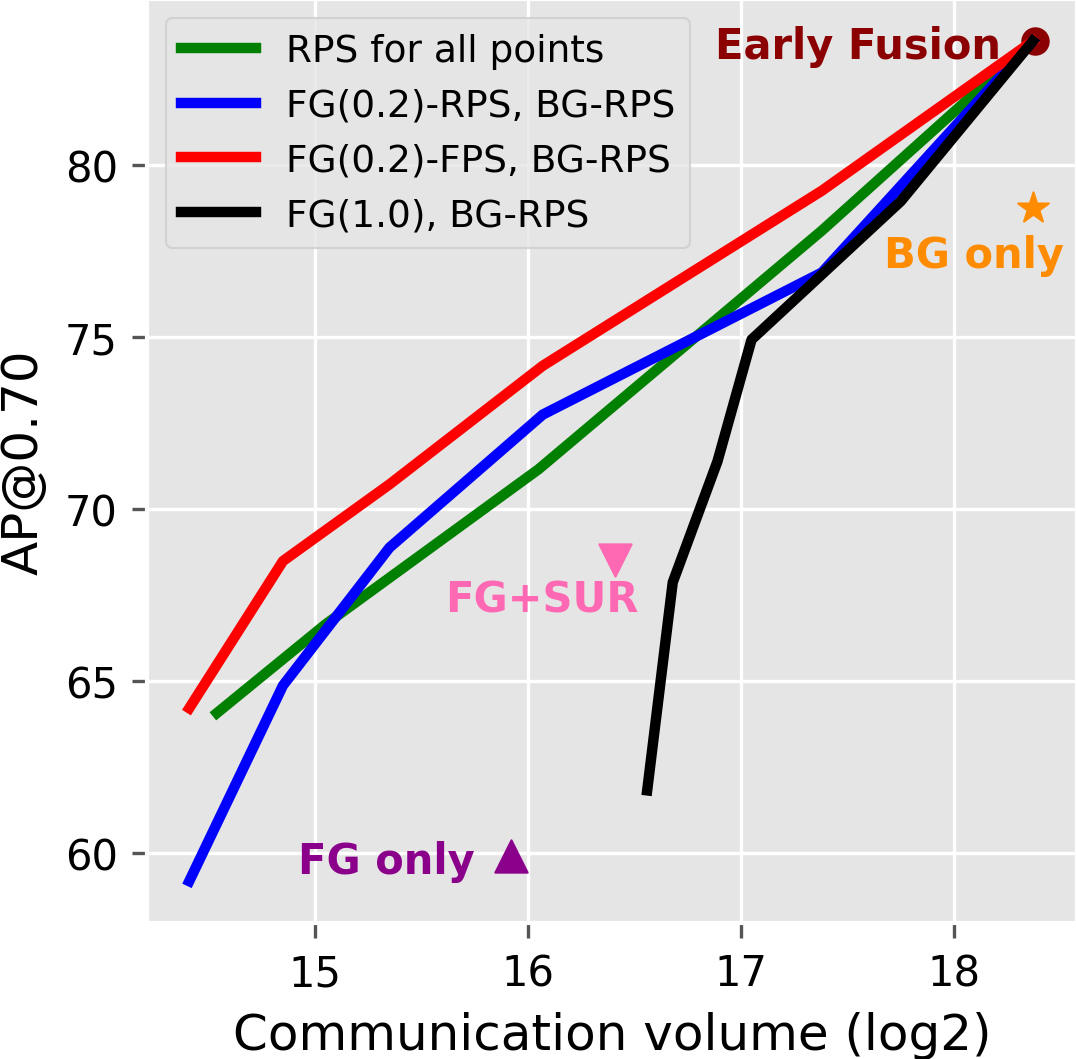}
        \captionsetup{font=small}
        \caption*{(f) Ablation on FAPS.}
    \end{minipage}
    \begin{minipage}[t]{0.245\textwidth}
        \centering
        \includegraphics[width=\textwidth]{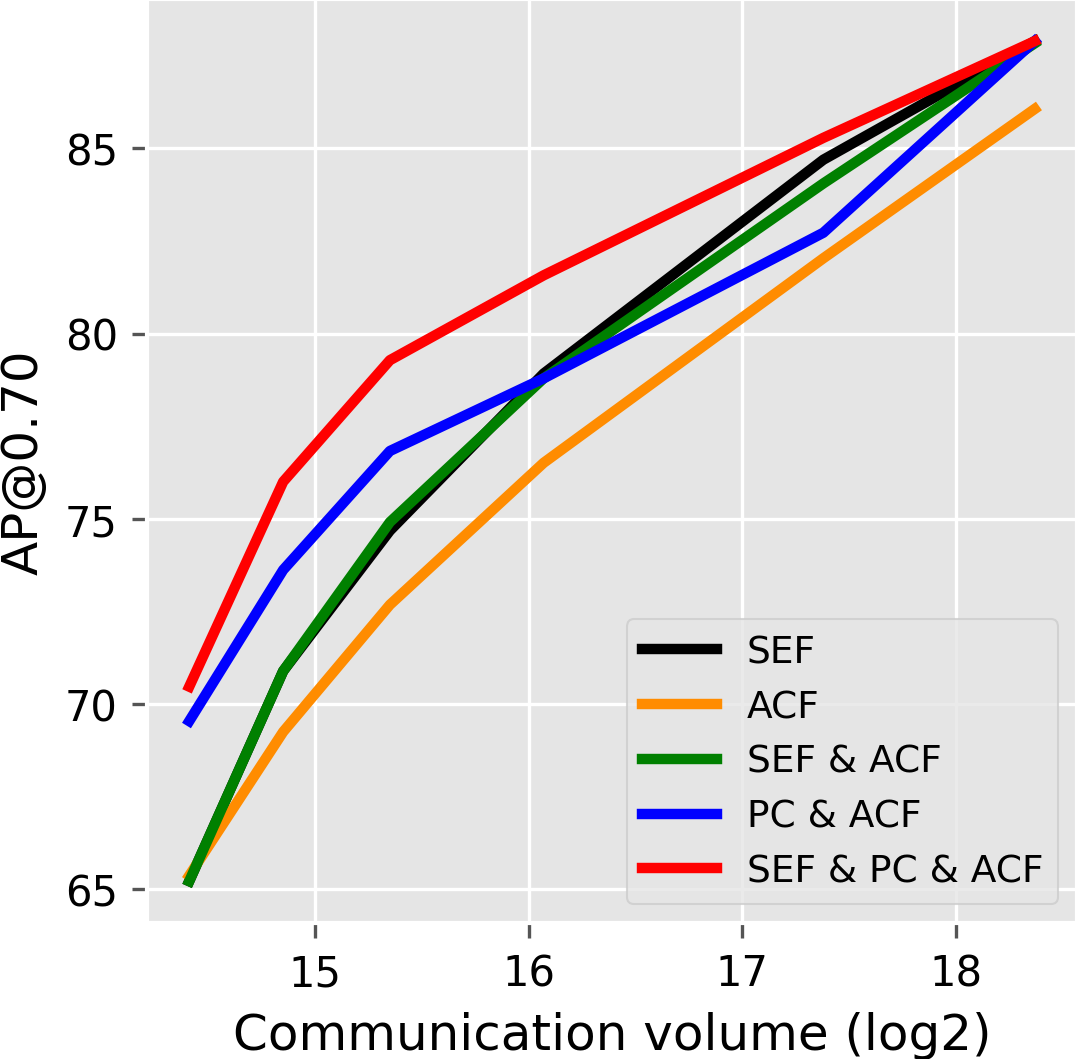}
        \captionsetup{font=small}
        \caption*{(g) Ablation on CEEF.}
    \end{minipage}
    \begin{minipage}[t]{0.245\textwidth}
        \centering
        \includegraphics[width=\textwidth]{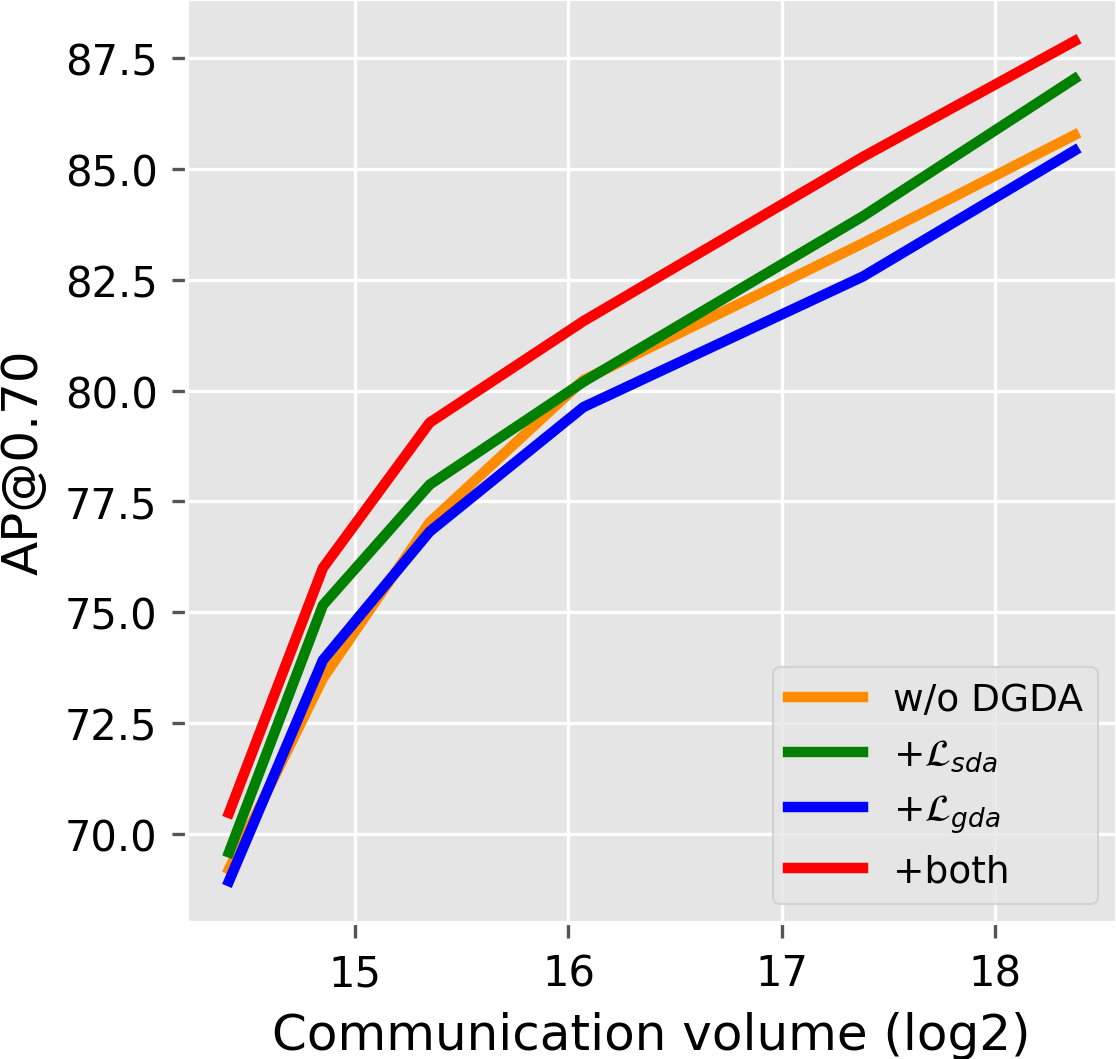}
        \captionsetup{font=small}
        \caption*{(h) Ablation on DGDA.}
    \end{minipage}
    \caption{Evaluation on V2XSim. In (b)-(d), CoLC transmits $100\%$ of the LiDAR data, while CoLC* transmits only $20\%$ of foreground points and $50\%$ of background points. In (e)-(h), when perform FAPS, $R^{fg}$ is fixed at 0.2, and $R^{bg}$ is varied from 0.5 to 0.01.}
    \label{fig5}
    \vspace{-0.5em}
\end{figure*}

\noindent
\textbf{Benchmark Comparison.} 
Table~\ref{tab1} compares collaborative detection performance across multiple datasets. 
On V2XSim, OPV2V, and V2XSet, CoLC yields the best AP@0.7 results, demonstrating the effectiveness of raw sensory data fusion in leveraging perceptual complementarity. 
Notably, when transmitting the full point cloud, it even slightly surpasses the early fusion baseline. 
This is because the completion and alignment during training enforce robust feature learning for CoLC, reducing overfitting to raw point clouds.
On DAIR-V2X, although CoLC attains slightly lower AP@0.7 than early fusion, it achieves higher AP@0.5, indicating stronger robustness under relaxed IoU thresholds in real-world scenarios.
Additionally, despite transmitting only half the communication volume, CoLC* maintains competitive performance, approaching or even surpassing the early-fusion baseline, demonstrating its robustness under communication constraints.

CoLC achieves an inference latency of 75.86 ms, remains comparable in speed to Where2comm (69.7ms), CoBEVT (84.5ms) and CoSDH (70.9 ms) and significantly faster than ERMVP (100.5 ms) and V2X-ViT (197.7 ms) on V2XSim.
 This communication efficiency ensures that agents are able to actively collaborate with each other.


\noindent
\textbf{Accuracy-Bandwidth Trade-Off Comparison.} 
Figure~\ref{fig5} (a) compares the proposed CoLC with prior methods in terms of the trade-off between detection performance and communication bandwidth. Numerical results are provided in the appendix. 
CoLC leverages FAPS to transmit sparse key points and LiDAR completion to restore information, thus achieving a superior perception-communication trade-off across all bandwidth levels.
Notably, although late fusion performs slightly better than CoLC under extremely low bandwidth, its performance lags significantly behind CoLC when sufficient bandwidth is available.

\noindent
\textbf{Robustness to Heterogeneous Scenarios.}
Figure~\ref{fig5} (b) further evaluates the robustness of the CoLC under heterogeneous scenarios. 
The homogeneous scenario refers to all agents using the same detector (e.g., PointPillars), whereas a heterogeneous scenario involves different models across agents (e.g., ego uses PointPillars while neighbors use separately trained SECOND \cite{second}). 
Intermediate and late fusion methods are susceptible to performance degradation in heterogeneous scenarios.
In extreme cases, intermediate fusion may even underperform the no fusion baseline.
In contrast, CoLC remains robust as it aggregates raw data from neighbor agents, avoiding semantic inconsistencies. 



\noindent
\textbf{Robustness to Localization Error and Communication Latency.}
Figure~\ref{fig5} (c) and (d) show the robustness of the models on pose error and transmission latency. The pose error setting follows CoAlign \cite{coalign} using Gaussian noise with a mean of 0m and standard deviations ranging from 0m to 1.0m. The latency setting follows V2X-ViT \cite{V2X-VIT}, varying from 0ms to 500ms. As the level of pose noise or latency increases, the perception performance of our CoLC degrades. 
However, it consistently outperforms the no fusion baseline under all perturbation levels and remains more robust than early fusion, whereas other intermediate fusion methods degrade significantly under large pose errors.

\subsection{Ablation Studies}

\paragraph{Ablation Studies on Three Components.}
Figure~\ref{fig5} (e) presents an ablation study on three key components. We begin by applying FAPS to the early fusion baseline, which reduces communication cost but degrades performance due to information loss. Introducing CEEF significantly improves performance by fusing completed dense pillars, suggesting that LiDAR completion can partially restore the lost information. Further incorporating DGDA leads to additional gains, indicating that DGDA effectively guides the adaptive fusion process. With all three components integrated, CoLC achieves a favorable balance between detection accuracy and communication efficiency. Compared to early fusion, CoLC achieves comparable performance with a $10\%$ reduction in communication cost, and yields a $5\%$ performance gain under the same communication volume.

\begin{figure*}[htbp]
    \centering
    \begin{minipage}[t]{0.16\textwidth}
        \centering
        \includegraphics[width=\textwidth]{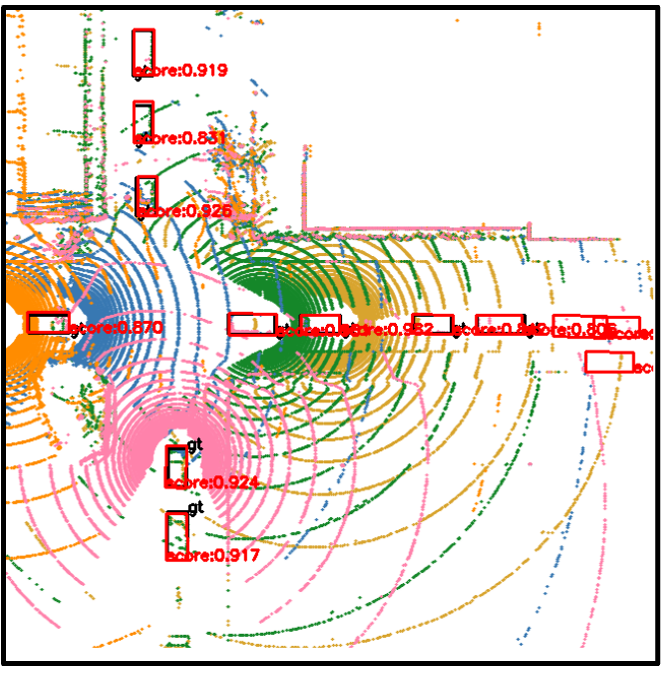}
        \caption*{(a) 100\% LiDAR}
    \end{minipage}
    \begin{minipage}[t]{0.16\textwidth}
        \centering
        \includegraphics[width=\textwidth]{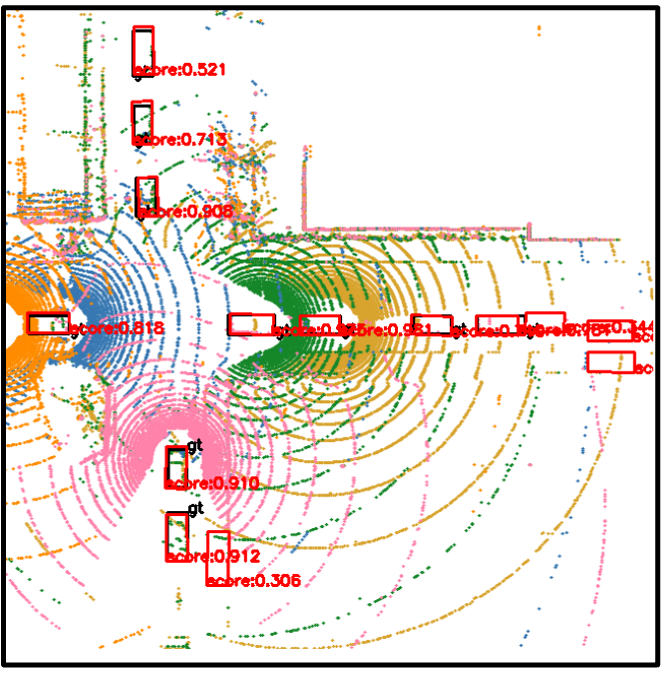}
        \caption*{(b) $R^{bg}=0.5$}
    \end{minipage}
    \begin{minipage}[t]{0.16\textwidth}
        \centering
        \includegraphics[width=\textwidth]{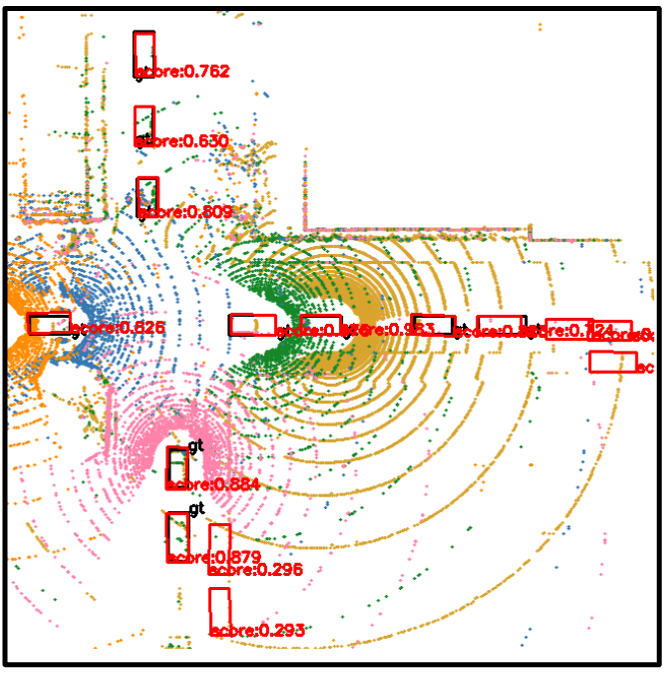}
        \caption*{(c) $R^{bg}=0.2$}
    \end{minipage}
    \begin{minipage}[t]{0.16\textwidth}
        \centering
        \includegraphics[width=\textwidth]{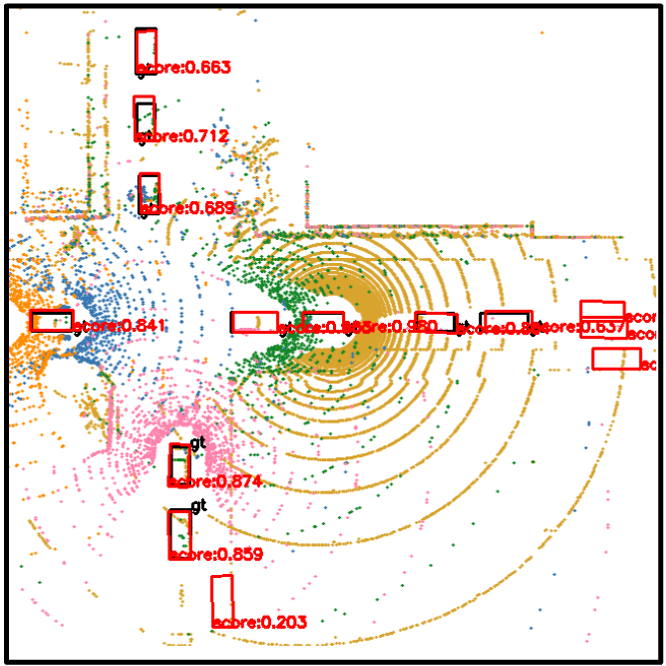}
        \caption*{(d) $R^{bg}=0.1$}
    \end{minipage}
    \begin{minipage}[t]{0.16\textwidth}
        \centering
        \includegraphics[width=\textwidth]{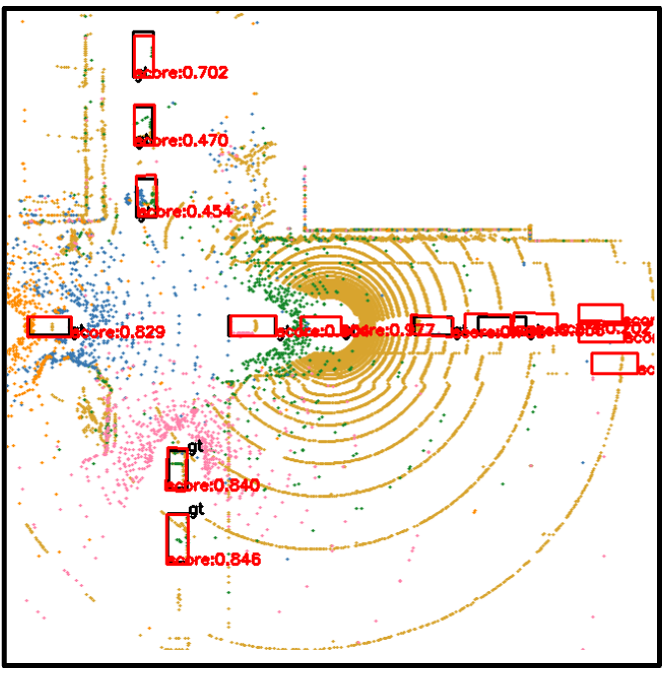}
        \caption*{(e) $R^{bg}=0.05$}
    \end{minipage}
    \begin{minipage}[t]{0.16\textwidth}
        \centering
        \includegraphics[width=\textwidth]{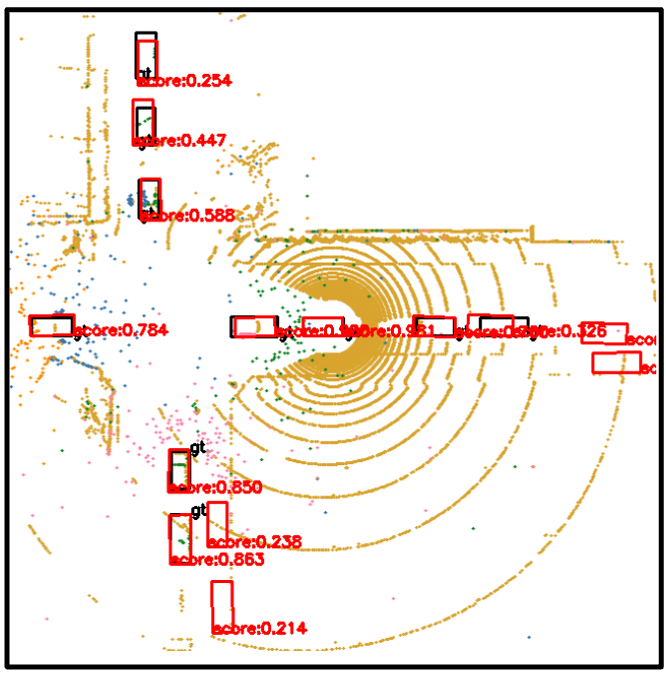}
        \caption*{(f) $R^{bg}=0.01$}
    \end{minipage}
    \caption{Detection results of CoLC under different point sampling ratios. 
    In (b)-(f), the foreground sampling ratio $R^{\text{fg}}$ is fixed at 0.2. 
    Black and \textcolor{red}{red} boxes represent ground truth and model predictions, respectively. 
    }
    \label{fig7}
    \vspace{-1.5em}
\end{figure*}

\noindent
\textbf{Ablation Studies on FAPS.}
Figure~\ref{fig5} (f) evaluates the effectiveness of FAPS under different foreground and background sampling strategies.
(1) Given a fixed foreground sampling ratio ($R^{fg}{=}0.2$), replacing FPS with RPS for foreground points degrades performance, indicating that FPS better preserves object structures.
(2) Under the same communication volume, applying RPS to all points performs worse than the hybrid setting (FG-FPS with $R^{fg}{=}0.2$, BG-RPS), confirming the benefit of structured sampling for foreground regions.
(3) Under the same communication volume, transmitting all foreground points in FAPS yields lower performance than transmitting only $20\%$ of them, indicating that a small subset of salient foreground points is sufficient for object representation, and more background regions provide essential contextual cues for accurate detection.
(4) Finally, our FAPS (FG-FPS with $R^{fg}{=}0.2$, BG-RPS) consistently outperforms transmitting foreground-only (FG only), foreground with surroundings (FG+SUR), or background-only (BG only) points, demonstrating that both foreground and background information are crucial for early fusion and FAPS preserves structural and contextual information.

\begin{table}[htbp]
    \caption{Ablation study on the codebook size ($K$), codebook dimension ($D_c$), sparse encoder dimension ($D$) and depth ($L$).
     }
    \label{tab4}
    \centering 
    \renewcommand\arraystretch{1.2} 
    \resizebox{1\linewidth}{!}{
    \begin{tabular}{cccc|cccc}
        \toprule
        $K$ & $D_c$ & $D$ & $L$ & IoU$\uparrow$ & MSE$\downarrow$& MB$\downarrow$ & AP@0.5/0.7$\uparrow$    \\
        \midrule
        128  & 128    & 128  & 3    &   0.473    &   0.060    & 39.08 &  88.14 / 75.93 \\
        128  & 128    & 128  & 6    &   0.585    &   0.052    & 44.07 &  \textbf{89.02} / 78.26 \\
        256  & 256    & 128   & 6      & 0.590          & 0.052         & 46.53 & 88.93 / 78.11 \\
    512  & 256    & 128   & 6     & 0.592          & 0.046          & 49.28 & 88.91 / 78.37  \\
    512  & 256    & 256  & 6     & 0.617          & 0.047          & 79.17 & 89.01 / 78.59   \\
    512  & 512    & 256   & 6   & \textbf{0.626} & \textbf{0.043} & 85.93 & 88.89 / \textbf{79.28}  \\
    1024 & 512    & 256   & 6     & 0.625          & 0.047          & 96.93 &  88.99 / 78.55  \\
    \bottomrule
    \end{tabular}
    }
  \end{table}
\begin{table}[htbp]
  \caption{Comparison of VQ-based and MAE-based completion.}
  \label{tab5}
  \centering 
  \renewcommand\arraystretch{1.1} 
  \resizebox{0.8\linewidth}{!}{
  \begin{tabular}{c|ccc}
      \toprule
     Method & IoU$\uparrow$ & MSE$\downarrow$& AP@0.5/0.7$\uparrow$  \\
      \midrule
  MAE-based  & \textbf{0.633}   & 0.057    & 88.17 / 77.55 \\
  VQ-based  & 0.626          & \textbf{0.043}   & \textbf{88.89 / 79.28} \\
  \bottomrule
  \end{tabular}
  }
  \vspace{-0.5em}
\end{table}


\noindent
\textbf{Ablation Studies on Progressive Fusion in CEEF.}
Figure~\ref{fig5} (g) ablates progressive fusion in CEEF: sparse early fusion (SEF), pillar completion (PC), and adaptive complementary fusion (ACF).
(1) Using SEF or ACF alone yields limited performance, though SEF consistently outperforms ACF since early fusion preserves more raw spatial information.
(2) At high communication volumes, combining SEF with ACF further improves performance because neighbor point clouds still contain rich information. In contrast, under low communication volumes, the combination of PC and ACF performs better, as LiDAR completion can effectively compensate for severe sparsity.
(3) Integrating all three stages (SEF, PC, and ACF) achieves the best performance across all bandwidth settings, demonstrating the complementary strengths of these fusion strategies.

\noindent
\textbf{Ablation Studies on Completion Module in CEEF.}
Table~\ref{tab4} ablates the VQ-based LiDAR completion module, evaluated with FAPS under $R^{fg}{=}0.2$ and $R^{bg}{=}0.1$. 
Increasing the codebook size and embedding dimension improves completion quality and increases the model size. In contrast, a lightweight configuration slightly reduces completion accuracy while halving the model size. Notably, the detection performance plateaus once the completion quality reaches a threshold (e.g., IoU $\geq$ 0.585, MSE $\leq$ 0.052). This indicates that the requirement of detection for a spatially complete context is satisfied at this reconstruction level.


    Since STAR \cite{li2023star} employs MAE-based reconstruction, we further compare MAE-based and VQ-based completion paradigms in Table~\ref{tab5}. 
While the MAE-based method achieves a higher IoU in occupancy prediction, its completed structures exhibit a larger MSE, indicating poorer reconstruction fidelity. In contrast, the VQ-based completion preserves discrete structural priors, leading to more discriminative pillars and superior detection accuracy. Therefore, we adopt the VQ-based completion in CEEF.

\begin{table}[htbp]
  \caption{Ablation on $\mathcal{L}_{\text{sda}}$ and $\mathcal{L}_{\text{gda}}$, where IoU and MSE quantify the similarity between $\hat{\mathcal{P}}_i^{\text{de}}$ and $\mathcal{P}_i^{\text{de}}$.}
  \label{tab6}
  \centering 
  \resizebox{0.75\linewidth}{!}{
  \begin{tabular}{cc|ccc}
      \toprule
     $\mathcal{L}_{\text{sda}}$ & $\mathcal{L}_{\text{gda}}$ & IoU$\uparrow$ & MSE$\downarrow$& AP@0.5/0.7$\uparrow$  \\
      \midrule
  &  &  0.765  & 0.415 &  88.48 / 77.03 \\
  \checkmark &  & 0.831   & 0.140 & 88.75 / 77.88 \\
  &  \checkmark & 0.836  & 0.129 & 88.16 / 76.82 \\
  \checkmark & \checkmark &  0.834  & 0.118  & \textbf{88.89 / 79.28} \\
  \bottomrule
  \end{tabular}
  }
  \vspace{-1em}
\end{table}

\noindent
\textbf{Ablation Studies on DGDA.}
Table~\ref{tab6} presents the ablation study of $\mathcal{L}_{\text{sda}}$ and $\mathcal{L}_{\text{gda}}$, evaluated with FAPS under $R^{fg}{=}0.2$ and $R^{bg}{=}0.1$. Compared with CEEF without DGDA, $\mathcal{L}_{\text{sda}}$ improves IoU by enhancing semantic consistency, while $\mathcal{L}_{\text{gda}}$ reduces MSE through structural regularization. Combining both achieves the balance between semantic completeness and geometric fidelity, yielding the highest detection accuracy. Figure~\ref{fig5} (h) further shows that combining both losses yields consistently higher detection accuracy across varying communication volumes.


\subsection{Qualitative Evaluation}



Figure~\ref{fig7} presents the detection results of CoLC under varying point sampling ratios.
As the ratio decreases, fewer points lead to noisier boxes and more false positives. Nevertheless, CoLC maintains correct classifications, demonstrating that FAPS effectively selects informative points and CEEF compensates for missing information.
\section{Conclusion}

We propose CoLC, a communication-efficient early collaborative perception framework that balances perception performance and bandwidth cost.
Specifically, FAPS enables spatially-aware point sampling to preserve key structural and contextual information under limited communication. CEEF reconstructs dense pillars from sparse inputs and performs adaptive fusion to enhance spatial completeness. During training, DGDA aligns the enhanced early-fusion pillars with dense supervision in semantic and geometric spaces.
Extensive experiments show that CoLC achieves superior accuracy under bandwidth constraints and remains robust in heterogeneous model scenarios.



{
    \small
    \bibliographystyle{ieeenat_fullname}
    \bibliography{references.bib}

@String(CVPR= {IEEE Conf. Comput. Vis. Pattern Recog.})

@String(ICCV= {Int. Conf. Comput. Vis.})

@String(ECCV= {Eur. Conf. Comput. Vis.})

@String(ICLR = {Int. Conf. Learn. Represent.})

@String(AAAI = {AAAI})

@String(CVPR  = {CVPR})

@String(ICCV  = {ICCV})

@String(ECCV  = {ECCV})

@String(ICLR  = {ICLR})

@article{han2023collaborative,
  author={Han, Yushan and Zhang, Hui and Li, Huifang and Jin, Yi and Lang, Congyan and Li, Yidong},
  journal={ITSM}, 
  title={Collaborative Perception in Autonomous Driving: Methods, Datasets, and Challenges}, 
  year={2023},
  volume={15},
  number={6},
  pages={131-151}}

@inproceedings{han2023ssc3od,
  title={{SSC3OD}: Sparsely supervised collaborative 3d object detection from lidar point clouds},
  author={Han, Yushan and Zhang, Hui and Zhang, Honglei and Li, Yidong},
  booktitle={SMC},
  pages={1360--1367},
  year={2023}
}

@inproceedings{han2025codts,
  title={{CoDTS}: Enhancing sparsely supervised collaborative perception with a dual teacher-student framework},
  author={Han, Yushan and Zhang, Hui and Zhang, Honglei and Wang, Jing and Li, Yidong},
  booktitle={AAAI},
  volume={39},
  number={3},
  pages={3366--3373},
  year={2025}
}

@article{han2025cods,
  title={{CoDS}: Enhancing Collaborative Perception in Heterogeneous Scenarios via Domain Separation},
  author={Han, Yushan and Zhang, Hui and Zhang, Honglei and Ding, Chuntao and Cao, Yuanzhouhan and Li, Yidong},
  journal={TMC},
  year={2026},
  volume={25},
  number={3}, 
  pages={4286--4299},
  publisher={IEEE}
}

@inproceedings{who2com,
  title={Who2com: Collaborative perception via learnable handshake communication},
  author={Liu, Yen-Cheng and Tian, Junjiao and Ma, Chih-Yao and Glaser, Nathan and Kuo, Chia-Wen and Kira, Zsolt},
  booktitle={ICRA},
  pages={6876--6883},
  year={2020}
}

@inproceedings{when2com,
  title={When2com: Multi-agent perception via communication graph grouping},
  author={Liu, Yen-Cheng and Tian, Junjiao and Glaser, Nathaniel and Kira, Zsolt},
  booktitle={CVPR},
  pages={4106--4115},
  year={2020}
}

@article{xia2026dota,
  title={{DOtA++}: Unsupervisely and Collaboratively Detect Objects From Multi-Agent Observations With Multi-Modal Prior Constraints},
  author={Xia, Qiming and Zheng, Longhui and Zhao, Shijia and Huang, Xun and Wu, Hai and Wen, Chenglu and Wang, Cheng},
  journal={TPAMI},
  year={2026},
  publisher={IEEE}
}

@InProceedings{ERMVP,
    author    = {Zhang, Jingyu and Yang, Kun and Wang, Yilei and Wang, Hanqi and Sun, Peng and Song, Liang},
    title     = {{ERMVP}: Communication-Efficient and Collaboration-Robust Multi-Vehicle Perception in Challenging Environments},
    booktitle = {CVPR},
    year      = {2024},
    pages     = {12575-12584}
}

@InProceedings{MRCNet,
    author    = {Hong, Shixin and Liu, Yu and Li, Zhi and Li, Shaohui and He, You},
    title     = {Multi-agent Collaborative Perception via Motion-aware Robust Communication Network},
    booktitle = {CVPR},
    year      = {2024},
    pages     = {15301-15310}
}

@inproceedings{cooper,
  title={{Cooper}: Cooperative perception for connected autonomous vehicles based on 3d point clouds},
  author={Chen, Qi and Tang, Sihai and Yang, Qing and Fu, Song},
  booktitle={ICDCS},
  pages={514--524},
  year={2019}
}

@article{arnold2020cooperative,
  title={Cooperative perception for 3D object detection in driving scenarios using infrastructure sensors},
  author={Arnold, Eduardo and Dianati, Mehrdad and De Temple, Robert and Fallah, Saber},
  journal={TITS},
  volume={23},
  number={3},
  pages={1852--1864},
  year={2020},
  publisher={IEEE}
}

@inproceedings{v2vnet,
  title={{V2VNet}: Vehicle-to-vehicle communication for joint perception and prediction},
  author={Wang, Tsun-Hsuan and Manivasagam, Sivabalan and Liang, Ming and Yang, Bin and Zeng, Wenyuan and Urtasun, Raquel},
  booktitle={ECCV},
  pages={605--621},
  year={2020}
}

@inproceedings{xu2023bridging,
  title={Bridging the domain gap for multi-agent perception},
  author={Xu, Runsheng and Li, Jinlong and Dong, Xiaoyu and Yu, Hongkai and Ma, Jiaqi},
  booktitle={ICRA},
  pages={6035--6042},
  year={2023}
}

@inproceedings{xu2023model,
  title={Model-agnostic multi-agent perception framework},
  author={Xu, Runsheng and Chen, Weizhe and Xiang, Hao and Xia, Xin and Liu, Lantao and Ma, Jiaqi},
  booktitle={ICRA},
  pages={1471--1478},
  year={2023}
}

@inproceedings{coalign,
  title={Robust collaborative 3d object detection in presence of pose errors},
  author={Lu, Yifan and Li, Quanhao and Liu, Baoan and Dianati, Mehrdad and Feng, Chen and Chen, Siheng and Wang, Yanfeng},
  booktitle={ICRA},
  pages={4812--4818},
  year={2023}
}

@InProceedings{OPV2V,
  author       = {Xu, Runsheng and Xiang, Hao and Xia, Xin and Han, Xu and Li, Jinlong and Ma, Jiaqi},
  booktitle    = {ICRA},
  title        = {{OPV2V}: An open benchmark dataset and fusion pipeline for perception with vehicle-to-vehicle communication},
  pages={2583--2589},
  year         = {2022},
  pages        = {2583--2589}
}

@inproceedings{DiscoNet,
  author    = {Li, Yiming and Ren, Shunli and Wu, Pengxiang and Chen, Siheng and Feng, Chen and Zhang, Wenjun},
  title     = {Learning distilled collaboration graph for multi-agent perception},
  year      = {2021},
  volume={34},
  pages={29541--29552},
  booktitle = {NeurIPS}
}

@inproceedings{where2comm,
  title={Where2comm: Communication-Efficient Collaborative Perception via Spatial Confidence Maps},
  author={Hu, Yue and Fang, Shaoheng and Lei, Zixing and Zhong, Yiqi and Chen, Siheng},
  booktitle={NeurIPS},
  year={2022},
  volume={35},
  pages={4874--4886},
}

@InProceedings{V2X-VIT,
  author  = {Xu, Runsheng and Xiang, Hao and Tu, Zhengzhong and Xia, Xin and Yang, Ming-Hsuan and Ma, Jiaqi},
  title   = {{V2X-ViT}: Vehicle-to-everything cooperative perception with vision transformer},
  booktitle={ECCV},
  pages={107--124},
  year={2022}
}

@inproceedings{xu2022cobevt,
title={Co{BEVT}: Cooperative Bird{\textquoteright}s Eye View Semantic Segmentation with Sparse Transformers},
author={Runsheng Xu and Zhengzhong Tu and Hao Xiang and Wei Shao and Bolei Zhou and Jiaqi Ma},
booktitle={CoRL},
year={2022},
}

@InProceedings{core,
    author    = {Wang, Binglu and Zhang, Lei and Wang, Zhaozhong and Zhao, Yongqiang and Zhou, Tianfei},
    title     = {{CORE}: Cooperative Reconstruction for Multi-Agent Perception},
    booktitle = {ICCV},
    year      = {2023},
    pages     = {8710-8720}
}

@inproceedings{yang2023how2comm,
  title={How2comm: Communication-efficient and collaboration-pragmatic multi-agent perception},
  author={Yang, Dingkang and Yang, Kun and Wang, Yuzheng and Liu, Jing and Xu, Zhi and Yin, Rongbin and Zhai, Peng and Zhang, Lihua},
  booktitle={NeurIPS},
  volume={36},
  year={2024}
}

@inproceedings{HEAL,
title={An Extensible Framework for Open Heterogeneous Collaborative Perception},
author={Lu, Yifan and Hu, Yue and Zhong, Yiqi and Wang, Dequan and Chen, Siheng and Wang, Yanfeng},
booktitle={ICLR},
year={2024},
}

@inproceedings{codefilling,
  title={Communication-efficient collaborative perception via information filling with codebook},
  author={Hu, Yue and Peng, Juntong and Liu, Sifei and Ge, Junhao and Liu, Si and Chen, Siheng},
  booktitle={CVPR},
  pages={15481--15490},
  year={2024}
}

@inproceedings{pnpda,
  title={Plug and Play: A Representation Enhanced Domain Adapter for Collaborative Perception},
  author={Luo, Tianyou and Yuan, Quan and Luo, Guiyang and Xia, Yuchen and Yang, Yujia and Li, Jinglin},
  booktitle={ECCV},
  pages={287--303},
  year={2024}
}

@inproceedings{pointcluster,
  title={Point Cluster: A Compact Message Unit for Communication-Efficient Collaborative Perception},
  author={Ding, Zihan and Fu, Jiahui and Liu, Si and Li, Hongyu and Chen, Siheng and Li, Hongsheng and Zhang, Shifeng and Zhou, Xu},
  booktitle={ICLR},
  year={2025}
}

@inproceedings{xu2025cosdh,
  title={CoSDH: Communication-Efficient Collaborative Perception via Supply-Demand Awareness and Intermediate-Late Hybridization},
  author={Xu, Junhao and Zhang, Yanan and Cai, Zhi and Huang, Di},
  booktitle={CVPR},
  pages={6834--6843},
  year={2025}
}

@inproceedings{polyinter,
  title={One is Plenty: A Polymorphic Feature Interpreter for Immutable Heterogeneous Collaborative Perception},
  author={Xia, Yuchen and Yuan, Quan and Luo, Guiyang and Fu, Xiaoyuan and Li, Yang and Zhu, Xuanhan and Luo, Tianyou and Chen, Siheng and Li, Jinglin},
  booktitle={CVPR},
  pages={1592--1601},
  year={2025}
}

@inproceedings{fu2025generative,
  title={Generative Map Priors for Collaborative BEV Semantic Segmentation},
  author={Fu, Jiahui and Gong, Yue and Wang, Luting and Zhang, Shifeng and Zhou, Xu and Liu, Si},
  booktitle={CVPR},
  pages={11919--11928},
  year={2025}
}

@inproceedings{yuan2025sparsealign,
  title={SparseAlign: A Fully Sparse Framework for Cooperative Object Detection},
  author={Yuan, Yunshuang and Xia, Yan and Cremers, Daniel and Sester, Monika},
  booktitle={CVPR},
  pages={22296--22305},
  year={2025}
}

@inproceedings{trafalign,
  title={{Traf-Align}: Trajectory-aware feature alignment for asynchronous multi-agent perception},
  author={Song, Zhiying and Yang, Lei and Wen, Fuxi and Li, Jun},
  booktitle={CVPR},
  pages={12048--12057},
  year={2025}
}

@inproceedings{li2023star,
  title={Multi-robot scene completion: Towards task-agnostic collaborative perception},
  author={Li, Yiming and Zhang, Juexiao and Ma, Dekun and Wang, Yue and Feng, Chen},
  booktitle={CoRL},
  pages={2062--2072},
  year={2023},
  organization={PMLR}
}

@inproceedings{yu2022dair,
  title={{DAIR-V2X}: A large-scale dataset for vehicle-infrastructure cooperative 3d object detection},
  author={Yu, Haibao and Luo, Yizhen and Shu, Mao and Huo, Yiyi and Yang, Zebang and Shi, Yifeng and Guo, Zhenglong and Li, Hanyu and Hu, Xing and Yuan, Jirui and others},
  booktitle={CVPR},
  pages={21361--21370},
  year={2022}
}

@inproceedings{pointpillars,
  title={{PointPillars}: Fast encoders for object detection from point clouds},
  author={Lang, Alex H and Vora, Sourabh and Caesar, Holger and Zhou, Lubing and Yang, Jiong and Beijbom, Oscar},
  booktitle={CVPR},
  pages={12697--12705},
  year={2019}
}

@article{second,
  title={{SECOND}: Sparsely embedded convolutional detection},
  author={Yan, Yan and Mao, Yuxing and Li, Bo},
  journal={Sensors},
  volume={18},
  number={10},
  pages={3337},
  year={2018}
}

@inproceedings{voxelnet,
  title={{VoxelNet}: End-to-end learning for point cloud based 3d object detection},
  author={Zhou, Yin and Tuzel, Oncel},
  booktitle={CVPR},
  pages={4490--4499},
  year={2018}
}

@inproceedings{liu2021swin,
  title={Swin {T}ransformer: Hierarchical vision transformer using shifted windows},
  author={Liu, Ze and Lin, Yutong and Cao, Yue and Hu, Han and Wei, Yixuan and Zhang, Zheng and Lin, Stephen and Guo, Baining},
  booktitle={ICCV},
  pages={10012--10022},
  year={2021}
}

@inproceedings{LiDARVAE,
  title={Deep generative modeling of lidar data},
  author={Caccia, Lucas and Van Hoof, Herke and Courville, Aaron and Pineau, Joelle},
  booktitle={IROS},
  pages={5034--5040},
  year={2019}
}

@inproceedings{ProjGAN,
  title={Projected gans converge faster},
  author={Sauer, Axel and Chitta, Kashyap and M{\"u}ller, Jens and Geiger, Andreas},
  booktitle={NeurIPS},
  volume={34},
  pages={17480--17492},
  year={2021}
}

@inproceedings{ultralidar,
  title={Learning compact representations for lidar completion and generation},
  author={Xiong, Yuwen and Ma, Wei-Chiu and Wang, Jingkang and Urtasun, Raquel},
  booktitle={CVPR},
  pages={1074--1083},
  year={2023}
}

@inproceedings{olidm,
  title={{OLiDM}: Object-aware LiDAR Diffusion Models for Autonomous Driving},
  author={Yan, Tianyi and Yin, Junbo and Lang, Xianpeng and Yang, Ruigang and Xu, Cheng-Zhong and Shen, Jianbing},
  booktitle={AAAI},
  volume={39},
  number={9},
  pages={9121--9129},
  year={2025}
}

@inproceedings{nunes2024scaling,
  title={Scaling diffusion models to real-world 3d lidar scene completion},
  author={Nunes, Lucas and Marcuzzi, Rodrigo and Mersch, Benedikt and Behley, Jens and Stachniss, Cyrill},
  booktitle={CVPR},
  pages={14770--14780},
  year={2024}
}

@InProceedings{vqvae,
  title={Neural discrete representation learning},
  author={Van Den Oord, Aaron and Vinyals, Oriol and others},
  booktitle={NeurIPS},
  volume={30},
  year={2017}
}

@InProceedings{CRCNet,
  author    = {Luo, Guiyang and Zhang, Hui and Yuan, Quan and Li, Jinglin},
  booktitle = {ACM MM},
  title     = {Complementarity-Enhanced and Redundancy-Minimized Collaboration Network for Multi-agent Perception},
  pages={3578--3586},
  year      = {2022}
}

@inproceedings{pointnet++,
  title={Pointnet++: Deep hierarchical feature learning on point sets in a metric space},
  author={Qi, Charles Ruizhongtai and Yi, Li and Su, Hao and Guibas, Leonidas J},
  booktitle={NeurIPS},
  volume={30},
  year={2017}
}

@article{zhang2021fast,
  title={Fast and robust iterative closest point},
  author={Zhang, Juyong and Yao, Yuxin and Deng, Bailin},
  journal={TPAMI},
  volume={44},
  number={7},
  pages={3450--3466},
  year={2021}
}
}


\end{document}